\documentclass[10pt,twocolumn,letterpaper]{article}

\usepackage{cvpr}              

\usepackage[table, dvipsnames]{xcolor}
\usepackage{times}
\usepackage{epsfig}
\usepackage{graphicx}
\usepackage{amsmath}
\usepackage{amssymb}
\usepackage{subfiles}
\usepackage[accsupp]{axessibility}
\usepackage{makecell} 
\usepackage{wrapfig} 
\usepackage{graphicx}
\usepackage[utf8]{inputenc} 
\usepackage[T1]{fontenc}    
\usepackage{url}            
\usepackage{booktabs}       
\usepackage{amsfonts}       
\usepackage{nicefrac}       
\usepackage{microtype}      
\usepackage{bm}

\usepackage{array}
\usepackage{multirow}

\newlength\savewidth

\usepackage{xspace}

\newcommand{\diff}[1]{\color{NavyBlue}\footnotesize#1}

\usepackage{xspace}

\makeatletter
\DeclareRobustCommand\onedot{\futurelet\@let@token\@onedot}
\def\@onedot{\ifx\@let@token.\else.\null\fi\xspace}

\def\eg{\emph{e.g}\onedot} 
\def\ie{\emph{i.e}\onedot} 
 
\def\etc{\emph{etc}\onedot} 
 
\def\etal{\emph{et al}\onedot}
\makeatother

\newcolumntype{S}{>{\centering\arraybackslash}m{0.9cm}}
\newcolumntype{M}{>{\centering\arraybackslash}m{1.2cm}}
\newcolumntype{L}{>{\centering\arraybackslash}m{1.4cm}}
\definecolor{mygray}{gray}{.95}
\definecolor{mylightergray}{gray}{.99}
\definecolor{mygreen}{RGB}{10, 179, 33}
\usepackage{caption} 
\renewcommand\arraystretch{1.15}
\makeatletter
\newcommand{\thickhline}{%
    \noalign {\ifnum 0=`}\fi \hrule height 1pt
    \futurelet \reserved@a \@xhline
}
\newcolumntype{"}{@{\vrule width 1pt}}

\definecolor{mygray}{gray}{.95}
\definecolor{mylightergray}{gray}{.99}
\definecolor{mygreen}{RGB}{10, 179, 33}

\usepackage{pifont}
\newcommand{\cmark}{\text{\ding{51}}}
\newcommand{\xmark}{\text{\ding{55}}}

\usepackage{hyperref}
\hypersetup{colorlinks,breaklinks}

\usepackage[capitalize]{cleveref}
\crefname{section}{Sec.}{Secs.}
\Crefname{section}{Section}{Sections}
\Crefname{table}{Table}{Tables}
\crefname{table}{Tab.}{Tabs.}

\def\confName{CVPR}
\def\confYear{2023}

\begin{document}

\title{3D Human Mesh Estimation from Virtual Markers}

\author{Xiaoxuan Ma\textsuperscript{1} \quad Jiajun Su\textsuperscript{1} 
\quad Chunyu Wang \textsuperscript{3 \thanks{Corresponding author}}  \quad Wentao Zhu\textsuperscript{1} \quad 
 Yizhou Wang\textsuperscript{1, 2, 4} \\[1.5ex]
    \textsuperscript{1~}School of Computer Science, Center on Frontiers of Computing Studies, Peking University \\
    \textsuperscript{2~}Inst. for Artificial Intelligence, Peking University\\
    \textsuperscript{3~}Microsoft Research Asia\\
    \textsuperscript{4~}Nat'l Eng. Research Center of Visual Technology\\[1.1ex]
{\tt\small \{maxiaoxuan, sujiajun, wtzhu, yizhou.wang\}@pku.edu.cn, chnuwa@microsoft.com}\\	
}
\maketitle

\begin{abstract}

Inspired by the success of volumetric 3D pose estimation, some recent human mesh estimators propose to estimate 3D skeletons as intermediate representations, from which, the dense 3D meshes are regressed by exploiting the mesh topology. 
However, body shape information is lost in extracting skeletons, leading to mediocre performance. The advanced motion capture systems solve the problem by placing dense physical markers on the body surface, which allows to extract realistic meshes from their non-rigid motions. However, they cannot be applied to wild images without markers. In this work, we present an intermediate representation, named virtual markers, which learns 64 landmark keypoints on the body surface based on the large-scale mocap data in a generative style, mimicking the effects of physical markers. The virtual markers can be accurately detected from wild images and can reconstruct the intact meshes with realistic shapes by simple interpolation. Our approach outperforms the state-of-the-art methods on three datasets. In particular, it surpasses the existing methods by a notable margin on the SURREAL dataset, which has diverse body shapes. Code is available at \url{https://github.com/ShirleyMaxx/VirtualMarker}.

\end{abstract}

\section{Introduction}
\label{sec:intro}

\begin{figure}[t]
    \centering
    \includegraphics[width=3in]{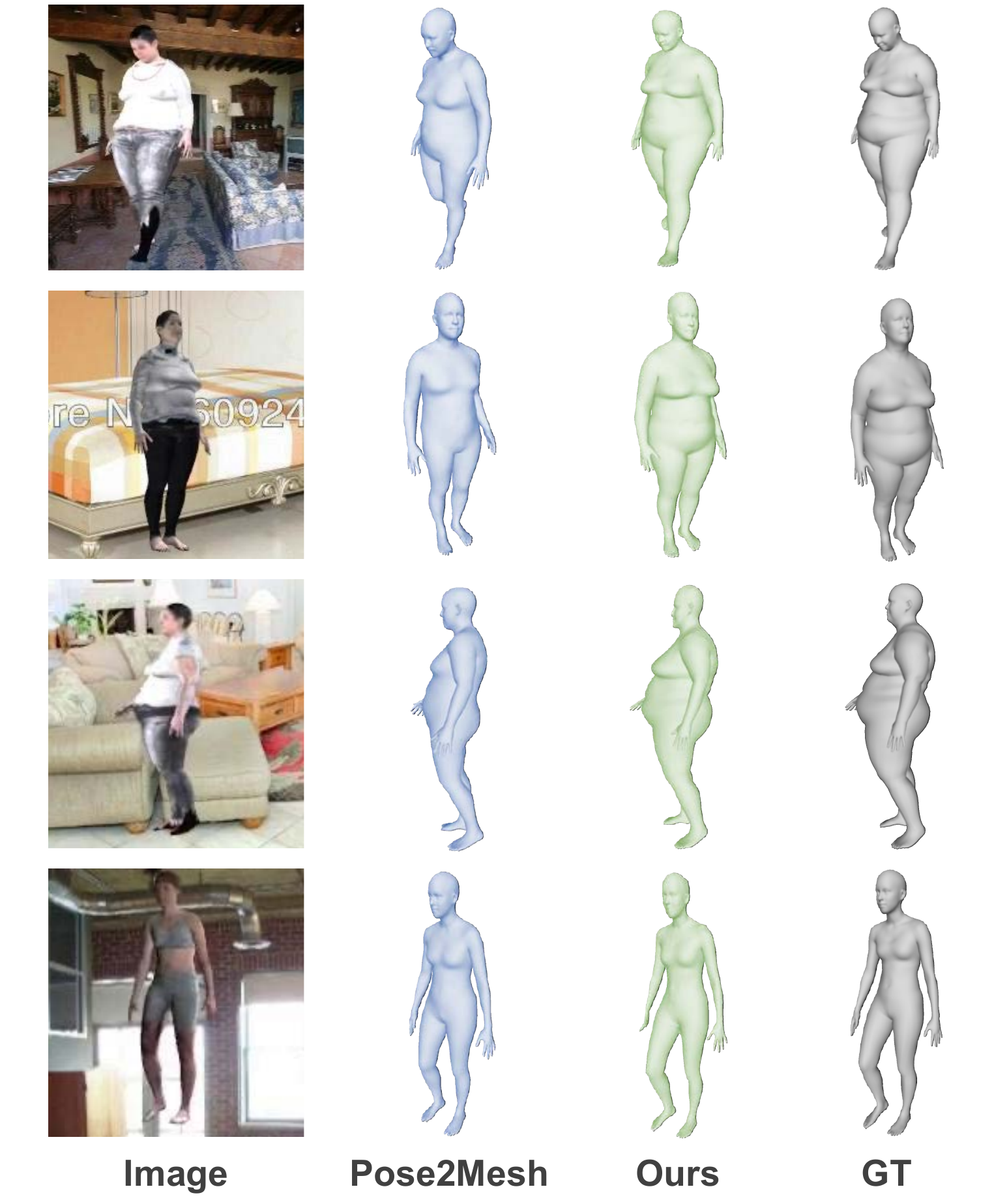}
    \caption{Mesh estimation results on four examples with different body shapes. Pose2Mesh \cite{choi2020pose2mesh} which uses 3D skeletons as the intermediate representation fails to predict accurate shapes. Our virtual marker-based method obtains accurate estimates.}
    \label{fig:surreal_shape}
    \vspace{-0.5cm}
\end{figure}

3D human mesh estimation aims to estimate the 3D positions of the mesh vertices that are on the body surface. The task has attracted a lot of attention from the computer vision and computer graphics communities \cite{boulic1997integration, presti20163d, li2020pastanet,loper2014mosh,NIPS2017_ab452534, kanazawa2018end, kolotouros2019learning, luan2021pc, Kolotouros_2021_ICCV, ci2022gfpose} because it can benefit many applications such as virtual reality \cite{huang2017towards}. Recently, the deep learning-based methods \cite{kanazawa2018end,choi2020pose2mesh,li2021hybrik} have significantly advanced the accuracy on the benchmark datasets.

The pioneer methods \cite{NIPS2017_ab452534, kanazawa2018end} propose to regress the pose and shape parameters of the mesh models such as SMPL \cite{loper2015smpl} directly from images. While straightforward, their accuracy is usually lower than the state-of-the-arts. The first reason is that the mapping from the image features to the model parameters is highly non-linear and suffers from image-model misalignment \cite{li2021hybrik}. Besides, existing mesh datasets \cite{h36m_pami,vonMarcard2018,mehta2017monocular,lassner2017unite} are small and limited to simple laboratory environments due to the complex capturing process. The lack of sufficient training data severely limits its performance.

Recently, some works \cite{kolotouros2019convolutional, moon2020i2l} begin to formulate mesh estimation as a dense 3D keypoint detection task inspired by the success of volumetric pose estimation \cite{sun2018integral,tu2020voxelpose,zhang2022voxeltrack,qiu2019cross,ye2022faster,su2022virtualpose}. For example, in \cite{kolotouros2019convolutional, moon2020i2l}, the authors propose to regress the 3D positions of all vertices. However, it is computationally expensive because it has more than several thousand vertices. Moon and Lee \cite{moon2020i2l} improve the efficiency by decomposing the 3D heatmaps into multiple 1D heatmaps at the cost of mediocre accuracy. Choi \etal \cite{choi2020pose2mesh} propose to first detect a sparser set of skeleton joints in the images,
from which the dense 3D meshes are regressed by exploiting the mesh topology. The methods along this direction have attracted increasing attention \cite{choi2020pose2mesh,li2021hybrik, wan2021encoder} due to two reasons. First, the proxy task of 3D skeleton estimation can leverage the abundant 2D pose datasets which notably improves the accuracy. Second, mesh regression from the skeletons is efficient. However, important information about the body shapes is lost in extracting the 3D skeletons, which is largely overlooked previously.  As a result, different types of body shapes, such as lean or obese, cannot be accurately estimated (see Figure \ref{fig:surreal_shape}).

The professional marker-based motion capture (mocap) method MoSh \cite{loper2014mosh} places physical markers on the body surface and explore their subtle non-rigid motions to extract meshes with accurate shapes. However, the physical markers limit the approach to be used in laboratory environments. We are inspired to think whether we can identify a set of landmarks on the mesh as virtual markers, \eg, elbow and wrist, that can be detected from wild images, and allow to recover accurate body shapes? The desired virtual markers should satisfy several requirements. First, the number of markers should be much smaller than that of the mesh vertices so that we can use volumetric representations to efficiently estimate their 3D positions. Second, the markers should capture the mesh topology so that the intact mesh can be accurately regressed from them. Third, the virtual markers have distinguishable visual patterns so that they can be detected from images.

In this work, we present a learning algorithm based on archetypal analysis \cite{cutler1994archetypal} to identify a subset of mesh vertices as the virtual markers that try to satisfy the above requirements to the best extent. Figure \ref{fig:body_arche} shows that the learned virtual markers coarsely outline the body shape and pose which paves the way for estimating meshes with accurate shapes. Then we present a simple framework for 3D mesh estimation on top of the representation as shown in Figure \ref{fig:pipeline}. It first learns a 3D keypoint estimation network based on \cite{sun2018integral} to detect the 3D positions of the virtual markers. Then we recover the intact mesh simply by interpolating them. The interpolation weights are pre-trained in the representation learning step and will be adjusted by a light network based on the prediction confidences of the virtual markers for each image. 

We extensively evaluate our approach on three benchmark datasets. It consistently outperforms the state-of-the-art methods on all of them. In particular, it achieves a significant gain on the SURREAL dataset \cite{varol2017learning} which has a variety of body shapes. Our ablation study also validates the advantages of the virtual marker representation in terms of recovering accurate shapes. Finally, the method shows decent generalization ability and generates visually appealing results for the wild images.

\section{Related work}
\label{sec:related}

\subsection{Optimization-based mesh estimation}
Before deep learning dominates this field, 3D human mesh estimation \cite{loper2014mosh,bogo2016keep, lassner2017unite, pavlakos2019expressive, zanfir2018monocular}  is mainly optimization-based, which optimizes the parameters of the human mesh models to match the observations. For example, Loper \etal \cite{loper2014mosh} propose MoSh that optimizes the SMPL parameters to align the mesh with the 3D marker positions. It is usually used to get GT 3D meshes for benchmark datasets because of its high accuracy. Later works propose to optimize the model parameters or mesh vertices based on 2D image cues \cite{bogo2016keep, lassner2017unite, pavlakos2019expressive, zanfir2018monocular, corona2022learned}. They extract intermediate representations such as 2D skeletons from the images and optimize the mesh model by minimizing the discrepancy between the model projection and the intermediate representations such as the 2D skeletons. These methods are usually sensitive to initialization and suffer from local optimum.

\begin{figure*}[t]
    \centering
    \includegraphics[width=6.3in]{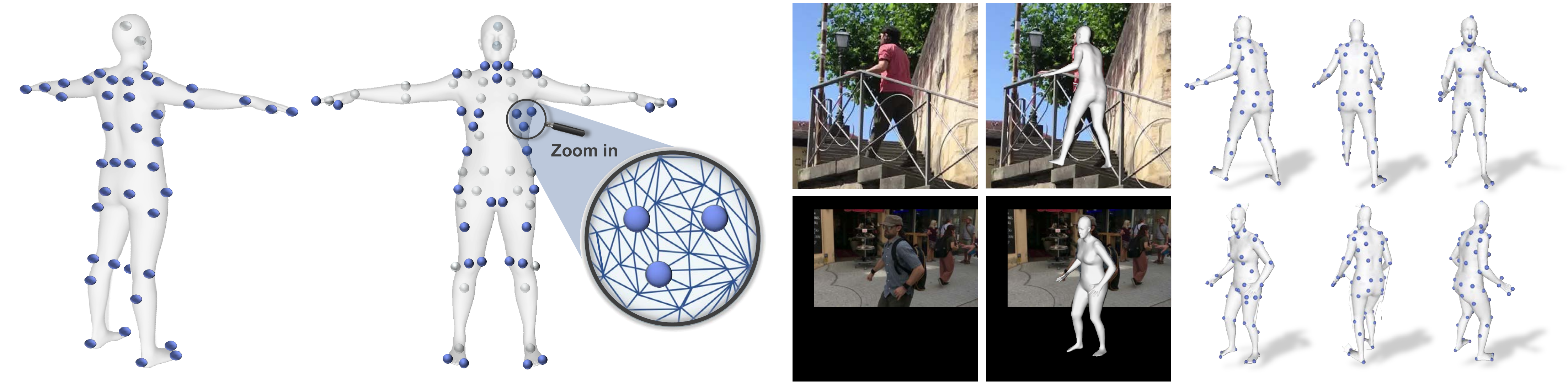}
    \caption{\textbf{Left:} The learned virtual markers (blue balls) in the back and front views. The grey balls mean they are invisible in the front view. The virtual markers act similarly to physical body markers and approximately outline the body shape. \textbf{Right:} Mesh estimation results by our approach, from left to right are input image, estimated 3D mesh overlayed on the image, and three different viewpoints showing the estimated 3D mesh with our intermediate predicted virtual markers (blue balls), respectively. }
    \label{fig:body_arche}
    \vspace{-0.4cm}
\end{figure*}

\subsection{Learning-based mesh estimation}
Recently, most works follow the learning-based framework and have achieved promising results. Deep networks \cite{NIPS2017_ab452534, kanazawa2018end, kolotouros2019learning, luan2021pc, Kolotouros_2021_ICCV} are used to regress the SMPL parameters from image features. However, learning the mapping from the image space to the parameter space is highly non-linear \cite{moon2020i2l}. In addition, they suffer from the misalignment between the meshes and image pixels \cite{zeng20203d}. These problems make it difficult to learn an accurate yet generalizable model. 

Some works propose to introduce proxy tasks to get intermediate representations first, hoping to alleviate the learning difficulty. In particular, intermediate representations of physical markers \cite{zanfir2021thundr}, IUV images \cite{xu2019denserac, zeng20203d, zhang2021pymaf, zhang2020densepose2smpl}, body part segmentation masks \cite{varol2018bodynet, Kocabas_2021_ICCV, lassner2017unite, omran2018neural} and body skeletons \cite{sun2019human, choi2020pose2mesh, li2021hybrik, wan2021encoder} have been proposed. In particular, THUNDR \cite{zanfir2021thundr} first estimates the 3D locations of physical markers from images and then reconstructs the mesh from the 3D markers. The physical markers can be interpreted as a simplified representation of body shape and pose. Although it is very accurate, it cannot be applied to wild images without markers. In contrast, body skeleton is a popular human representation that can be robustly detected from wild images. Choi \etal \cite{choi2020pose2mesh} propose to first estimate the 3D skeletons, and then estimate the intact mesh from them. However, accurate body shapes are difficult to be recovered from the oversimplified 3D skeletons.

Our work belongs to the learning-based class and is related to works that use physical markers or skeletons as intermediate representations. But different from them, we propose a novel intermediate representation, named \textit{virtual markers}, which is more expressive to reduce the ambiguity in pose and shape estimation than body skeletons and can be applied to wild images.

\section{Method}

\label{sec:method}

In this section, we describe the details of our approach. First, Section \ref{subsec:body_arche} introduces how we learn the virtual marker representation from mocap data. Then we present the overall framework for mesh estimation from an image in Section \ref{subsec:approach}. At last, Section \ref{subsec:training} discusses the loss functions and training details.

\subsection{The virtual marker representation} 
\label{subsec:body_arche}

We represent a mesh by a vector of vertex positions $\mathbf{x} \in \mathbb{R}^{3M}$ where $M$ is the number of mesh vertices. Denote a mocap dataset such as \cite{h36m_pami} with $N$ meshes as $\overset{\frown}{\mathbf{X}} = [\mathbf{x}_1 ,\, ... ,\, \mathbf{x}_N] \in \mathbb{R}^{3M \times N}$. To unveil the latent structure among vertices, we reshape it to $\mathbf{X} \in \mathbb{R}^{3N \times M}$ with each column $\mathbf{x}_i \in \mathbb{R}^{3N}$ representing all possible positions of the $i^{\text{th}}$ vertex in the dataset \cite{h36m_pami}.

The rank of $\mathbf{X}$ is smaller than $M$ because the mesh representation is smooth and redundant where some vertices can be accurately reconstructed by the others. While it seems natural to apply PCA \cite{jolliffe1986principal} to $\mathbf{X}$ to compute the eigenvectors as virtual markers for reconstructing others, there is no guarantee that the virtual markers correspond to the mesh vertices, making them difficult to be detected from images. Instead, we aim to learn $K$ virtual markers $\mathbf{Z} = [\mathbf{z}_1, ..., \mathbf{z}_K] \in \mathbb{R}^{3N \times K}$ that try to satisfy the following two requirements to the greatest extent. First, they can accurately reconstruct the intact mesh $\mathbf{X}$ by their linear combinations: $\mathbf{X}=\mathbf{Z} \mathbf{A}$, where $\mathbf{A} \in \mathbb{R}^{K \times M}$ is a coefficient matrix that encodes the spatial relationship between the virtual markers and the mesh vertices. Second, they should have distinguishable visual patterns in images so that they can be easily detected from images. Ideally, they can be on the body surface as the meshes. 

We apply archetypal analysis \cite{cutler1994archetypal,chen2014fast} to learn $\mathbf{Z}$ by minimizing a reconstruction error with two additional constraints: (1) each vertex $\mathbf{x}_i$ can be reconstructed by convex combinations of $\mathbf{Z}$, and (2) each marker $\mathbf{z}_i$ should be convex combinations of the mesh vertices $\mathbf{X}$: 
\begin{equation}
\label{eq:aa}
\min_{\substack{\bm{\alpha}_i \in {\Delta}_K \, for \, 1\leq i\leq M, \\ 
                \bm{\beta}_j \in {\Delta}_M \, for \, 1\leq j\leq K}} ||\mathbf{X} - \mathbf{X}\mathbf{B}\mathbf{A}||^2_F, \\
\end{equation}
where $\mathbf{A} = [\bm{\alpha}_1, ..., \bm{\alpha}_M] \in \mathbb{R}^{K \times M}$, each $\bm{\alpha}$ resides in the simplex ${\Delta}_K \triangleq \{\bm{\alpha} \in \mathbb{R}^{K} \, \mathrm{s.t.} \, \bm{\alpha} \succeq 0 \, \text{and} \, {||\bm{\alpha}||}_1 = 1\}$, and $\mathbf{B} = [\bm{\beta}_1, ..., \bm{\beta}_K] \in \mathbb{R}^{M \times K}$, $\bm{\beta}_j \in {\Delta}_M$. We adopt Active-set algorithm \cite{chen2014fast} to solve objective (\ref{eq:aa}) and obtain the learned virtual markers $\mathbf{Z} = \mathbf{X}\mathbf{B} \in \mathbb{R}^{3N \times K}$.  As shown in \cite{cutler1994archetypal,chen2014fast}, the two constraints encourage the virtual markers $\mathbf{Z}$ to unveil the latent structure among vertices, therefore they learn to be close to the extreme points of the mesh and located on the body surface as much as possible. \\

\begin{figure*}[t]
	\centering
	\includegraphics[width=6.3in]{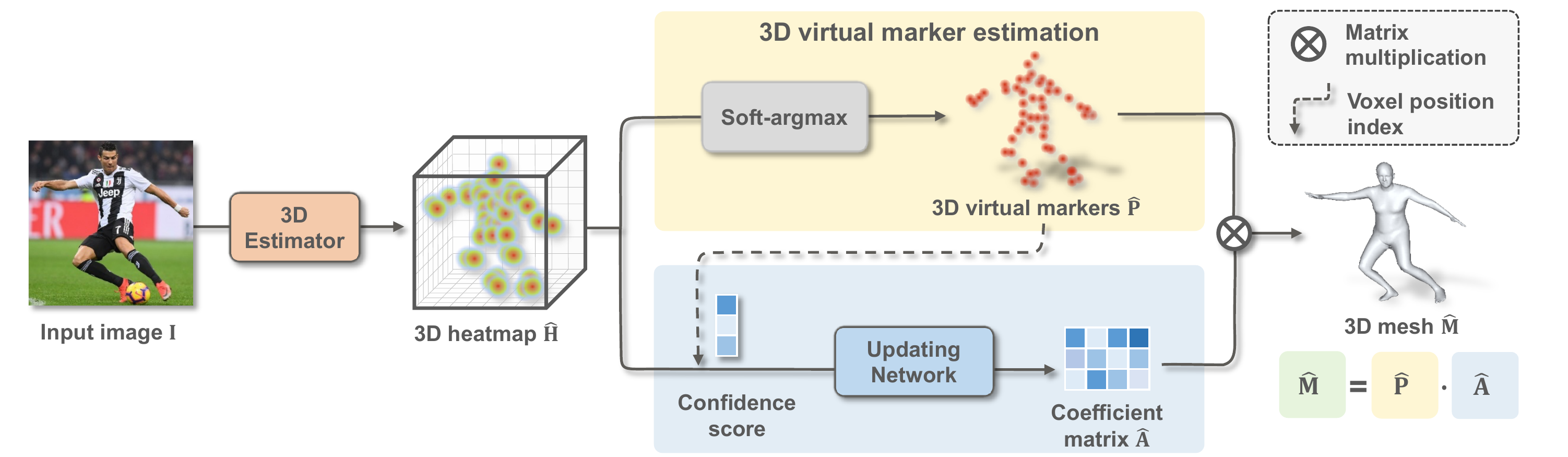}
	\caption{Overview of our framework. Given an input image $\mathbf{I}$, it first estimates the 3D positions $\hat{\mathbf{P}}$ of the virtual markers. Then we update the coefficient matrix $\hat{\mathbf{A}}$ based on the estimation confidence scores $\mathbf{C}$ of the virtual markers. Finally, the complete human mesh can be simply recovered by linear multiplication $\hat{\mathbf{M}} = \hat{\mathbf{P}}\hat{\mathbf{A}}$. }
	\label{fig:pipeline}
    \vspace{-0.4cm}
\end{figure*}

\begin{table}[t]
    \centering
    \resizebox{3.0in}{!}{
    \begin{tabular}{l|l|c} 
    \hline 
    Type & Formula & Reconst. Error (mm) $\downarrow$  \\
    \hline  
    Original & $||\mathbf{X} - \mathbf{X}\mathbf{B}\mathbf{A}||^2_F$ & 11.67 \\
    Symmetric & $||\mathbf{X} - \mathbf{X}\widetilde{\mathbf{B}}^{sym}\widetilde{\mathbf{A}}^{sym}||^2_F$ & 10.98 \\
    \hline
    \end{tabular}}
    \caption{The reconstruction errors using the original and the symmetric sets of markers on the H3.6M dataset \cite{h36m_pami}, respectively. The errors are small indicating that they are sufficiently expressive and can reconstruct all vertices accurately. }
    \label{tab:comparison_sym_arche}
    \vspace{-0.4cm}
\end{table}

\noindent\textbf{Post-processing.}
Since human body is left-right symmetric, we adjust $\mathbf{Z}$ to reflect the property. We first replace each $\mathbf{z}_i \in \mathbf{Z}$ by its nearest vertex on the mesh and obtain $\widetilde{\mathbf{Z}} \in \mathbb{R}^{3 \times K}$. This step allows us to compute the left or right counterpart of each marker. Then we replace the markers in the right body with the symmetric vertices in the left body and obtain the symmetric markers $\widetilde{\mathbf{Z}}^{sym} \in \mathbb{R}^{3 \times K}$. Finally we update $\mathbf{B}$ and $\mathbf{A}$ by minimizing $||\mathbf{X} - \mathbf{X}\widetilde{\mathbf{B}}^{sym}\widetilde{\mathbf{A}}^{sym}||^2_F$ subject to $\widetilde{\mathbf{Z}}^{sym} = \mathbf{X} \widetilde{\mathbf{B}}^{sym}$. 
More details are elaborated in the supplementary.

Figure \ref{fig:body_arche} shows the virtual markers learned on the mocap dataset \cite{h36m_pami} after post-processing. They are similar to the physical markers and approximately outline the body shape which agrees with our expectations. They are roughly evenly distributed on the surface of the body, and some of them are located close to the body keypoints, which have distinguishable visual patterns to be accurately detected. Table \ref{tab:comparison_sym_arche} shows the reconstruction errors of using original markers $\mathbf{X}\mathbf{B}$ and the symmetric markers $\mathbf{X}\widetilde{\mathbf{B}}^{sym}$. Both can reconstruct meshes accurately. 

\subsection{Mesh estimation framework}
\label{subsec:approach}

On top of the virtual markers, we present a simple yet effective framework for end-to-end 3D human mesh estimation from a single image. As shown in Figure \ref{fig:pipeline}, it consists of two branches. The first branch uses a volumetric CNN \cite{sun2018integral} to estimate the 3D positions $\hat{\mathbf{P}}$ of the markers, and the second branch reconstructs the full mesh $\hat{\mathbf{M}}$ by predicting a coefficient matrix $\hat{\mathbf{A}}$:
\vspace{-0.2cm}
\begin{equation}
\small
\hat{\mathbf{M}} = \hat{\mathbf{P}}\hat{\mathbf{A}}.
\end{equation}
We will describe the two branches in more detail. \\

\noindent\textbf{3D marker estimation.}
We train a neural network to estimate a 3D heatmap $\hat{\mathbf{H}} = [\hat{\mathbf{H}}_1 ,\, ... ,\, \hat{\mathbf{H}}_{K}] \in \mathbb{R}^{K \times D \times H \times W}$ from an image. The heatmap encodes per-voxel likelihood of each marker. There are $D \times H \times W$ voxels in total which are used to discretize the 3D space. The 3D position $\hat{\mathbf{P}}_z \in \mathbb{R}^{3}$ of each marker is computed as the center of mass of the corresponding heatmap $\hat{\mathbf{H}}_z$ \cite{sun2018integral} as follows:
\begin{equation}
\small
    \hat{\mathbf{P}}_z = \sum_{d=1}^{D}\sum_{h=1}^{H}\sum_{w=1}^{W}(d, h, w) \cdot \hat{\mathbf{H}}_z(d, h, w).
\end{equation}
The positions of all markers are represented as $\hat{\mathbf{P}} = [\hat{\mathbf{P}}_1, \hat{\mathbf{P}}_2, \cdots, \hat{\mathbf{P}}_K]$. \\

\noindent\textbf{Interpolation.} 
Ideally, if we have accurate estimates for all virtual markers $\hat{\mathbf{P}}$, then we can recover the complete mesh by simply multiplying $\hat{\mathbf{P}}$ with a fixed coefficient matrix $\widetilde{\mathbf{A}}^{sym}$ with sufficient accuracy as validated in Table \ref{tab:comparison_sym_arche}. However, in practice, some markers may have large estimation errors because they may be occluded in the monocular setting. Note that this happens frequently. For example, the markers in the back will be occluded when a person is facing the camera. As a result, inaccurate markers positions may bring large errors to the final mesh if we directly multiply them with the fixed matrix $\widetilde{\mathbf{A}}^{sym}$.

Our solution is to rely more on those accurately detected markers. To that end, we propose to update the coefficient matrix based on the estimation confidence scores of the markers. In practice, we simply take the heatmap score at the estimated positions of each marker, \ie $\hat{\mathbf{H}}_z(\hat{\mathbf{P}}_z)$, and feed them to a single fully-connected layer to obtain the coefficient matrix $\hat{\mathbf{A}}$. Then the mesh is reconstructed by $\hat{\mathbf{M}} = \hat{\mathbf{P}}\hat{\mathbf{A}}$.

\subsection{Training}
\label{subsec:training}
We train the whole network end-to-end in a supervised way. 
The overall loss function is defined as:
\begin{equation}
\small
\label{eq:total_loss}
\begin{aligned}
    \mathcal{L} = \lambda_{vm}\mathcal{L}_{vm} + \lambda_{c}\mathcal{L}_{conf} + \lambda_{m}\mathcal{L}_{mesh}.
\end{aligned}
\end{equation}

\noindent\textbf{Virtual marker loss.}
We define $\mathcal{L}_{vm}$ as the $L_1$ distance between the predicted 3D virtual markers $\hat{\mathbf{P}}$ and the GT $\hat{\mathbf{P}}^{*}$ as follows:
\begin{equation}
\small
\label{eq:loss_pose}
\begin{aligned}
    \mathcal{L}_{vm} = \| \hat{\mathbf{P}}-\hat{\mathbf{P}}^{*} \|_1.
\end{aligned}
\end{equation}
Note that it is easy to get GT markers $\hat{\mathbf{P}}^{*}$ from GT meshes as stated in Section \ref{subsec:body_arche} without additional manual annotations. \\

\noindent\textbf{Confidence loss.}
We also require that the 3D heatmaps have reasonable shapes, therefore, the heatmap score at the voxel containing the GT marker position $\hat{\mathbf{P}}_z^{*}$ should have the maximum value as in the previous work \cite{iskakov2019learnable}:
\begin{equation}
\small
\label{eq:loss_conf}
\begin{aligned}
    \mathcal{L}_{conf} = - \sum_{z=1}^{K} log(\hat{\mathbf{H}}_z(\hat{\mathbf{P}}_z^{*})).
\end{aligned}
\end{equation}

\noindent\textbf{Mesh loss.}
Following \cite{moon2020i2l}, we define $\mathcal{L}_{mesh}$ as a weighted sum of four losses:
\begin{equation}
\label{eq:mesh_loss}
\begin{aligned}
\small
    \mathcal{L}_{mesh} = \mathcal{L}_{vertex} + \mathcal{L}_{pose} + \mathcal{L}_{normal} + \lambda_{e}\mathcal{L}_{edge}.
\end{aligned}
\end{equation}
\begin{itemize}
    \item[--] \textbf{Vertex coordinate loss.} We adopt $L_1$ loss between predicted 3D mesh coordinates $\hat{\mathbf{M}}$ with GT mesh $\hat{\mathbf{M}}^{*}$ as:
        \begin{equation}
        \small
        \label{eq:loss_vertex}
        \begin{aligned}
            \mathcal{L}_{vertex} = \| \hat{\mathbf{M}}-\hat{\mathbf{M}}^{*} \|_1.
        \end{aligned}
        \end{equation}
    \item[--] \textbf{Pose loss.} We use $L_1$ loss between the 3D landmark joints regressed from mesh $\hat{\mathbf{M}}\mathcal{J}$ and the GT joints $\hat{\mathbf{J}}^{*}$ as:
        \begin{equation}
        \small
        \label{eq:loss_reg}
        \begin{aligned}
            \mathcal{L}_{pose} = \| \hat{\mathbf{M}}\mathcal{J}-\hat{\mathbf{J}}^{*} \|_1,
        \end{aligned}
        \end{equation}
    where $\mathcal{J} \in \mathbb{R}^{M \times J}$ is a pre-defined joint regression matrix in SMPL model \cite{bogo2016keep}.

    \item[--] \textbf{Surface losses.} To improve surface smoothness \cite{wang2018pixel2mesh}, we supervise the normal vector of a triangle face with GT normal vectors by $\mathcal{L}_{normal}$ and the edge length of the predicted mesh with GT length by $\mathcal{L}_{edge}$:
        \begin{equation}
        \small
        \label{eq:loss_surface}
        \begin{aligned}
            \small
            & \mathcal{L}_{normal} = \sum_{f} \sum_{\{i, j\} \subset f}  \left| \left< \frac{\hat{\mathbf{M}}_i - \hat{\mathbf{M}}_j}{\| \hat{\mathbf{M}}_i - \hat{\mathbf{M}}_j \|_2}, \hat{\mathbf{n}}_f^{*}\right> \right|, \\
            \small
            & \mathcal{L}_{edge} = \sum_{f} \sum_{\{i, j\} \subset f}  \left|\| \hat{\mathbf{M}}_i - \hat{\mathbf{M}}_j \|_2 - \| \hat{\mathbf{M}}_i^{*} - \hat{\mathbf{M}}_j^{*} \|_2 \right|.
        \end{aligned}
        \end{equation}
    where $f$ and $\hat{\mathbf{n}}_f^{*}$ denote a triangle face in the mesh and its GT unit normal vector, respectively. $\hat{\mathbf{M}}_i$ denote the $i^{th}$ vertex of $\hat{\mathbf{M}}$. $^{*}$ denotes GT.
\end{itemize}

\section{Experiments}

\label{sec:experiments}

\begin{table*}[ht]
\center
\small
\setlength{\tabcolsep}{7pt}
\resizebox{6.0in}{!}{
\begin{tabular}{l l | c c c | c c c}
    \hline 
    \multirow{2}{*}{Method} & Intermediate & \multicolumn{3}{c}{H3.6M} & \multicolumn{3}{|c}{3DPW} \\

    \cline{3-5} \cline{6-8}
    & Representation & MPVE$\downarrow$ & MPJPE$\downarrow$ & PA-MPJPE$\downarrow$ & MPVE$\downarrow$ & MPJPE$\downarrow$ & PA-MPJPE$\downarrow$ \\
    \hline 
    $^{\dagger}$ Arnab \etal \cite{arnab2019exploiting} CVPR'19  & 2D skeleton & - & 77.8 & 54.3 & - & - & 72.2 \\
    $^{\dagger}$ HMMR \cite{kanazawa2019learning} CVPR'19 & - & - & - & 56.9 & 139.3 & 116.5 & 72.6 \\
    $^{\dagger}$ DSD-SATN \cite{sun2019human} ICCV'19 & 3D skeleton & - & 59.1 & 42.4 & - & - & 69.5 \\
    $^{\dagger}$ VIBE \cite{kocabas2020vibe} CVPR'20 & - & - & 65.9 & 41.5 & 99.1 & 82.9 & 51.9 \\
    $^{\dagger}$ TCMR \cite{choi2021beyond} CVPR'21 & - & - & 62.3 & 41.1 & 102.9 & 86.5 & 52.7 \\
    $^{\dagger}$ MAED \cite{wan2021encoder} ICCV'21 & 3D skeleton & - & 56.3 & 38.7 & 92.6 & 79.1 & 45.7 \\
    \hline
    SMPLify \cite{bogo2016keep} ECCV'16 & 2D skeleton & - & - & 82.3 & - & - & - \\
    HMR \cite{kanazawa2018end} CVPR'18 & - & 96.1 & 88.0 & 56.8 & 152.7 & 130.0 & 81.3 \\
    GraphCMR \cite{kolotouros2019convolutional} CVPR'19 & 3D vertices & - & - & 50.1 & - & - & 70.2 \\
    SPIN \cite{kolotouros2019learning} ICCV'19 & - & - & - & 41.1 & 116.4 & 96.9 & 59.2 \\
    DenseRac \cite{xu2019denserac} ICCV'19 & IUV image & - & 76.8 & 48.0 & - & - & - \\
    DecoMR \cite{zeng20203d} CVPR'20 & IUV image & - & 60.6 & 39.3 & - & - & - \\
    ExPose \cite{choutas2020monocular} ECCV'20 & - & - & - & - & - & 93.4 & 60.7 \\
    Pose2Mesh \cite{choi2020pose2mesh} ECCV'20 & 3D skeleton & 85.3 & 64.9 & 46.3 & 106.3 & 88.9 & 58.3 \\
    I2L-MeshNet \cite{moon2020i2l} ECCV'20 & 3D vertices & 65.1 & 55.7 & 41.1 & 110.1 & 93.2 & 57.7 \\
    PC-HMR \cite{luan2021pc} AAAI'21 & 3D skeleton & - & - & - & 108.6 & 87.8 & 66.9  \\
    HybrIK \cite{li2021hybrik} CVPR'21 & 3D skeleton & 65.7 & 54.4 & 34.5 & 86.5 & 74.1 & 45.0  \\
    METRO \cite{lin2021end} CVPR'21 & 3D vertices & - & 54.0 & 36.7 & 88.2 & 77.1 & 47.9 \\
    ROMP \cite{sun2021monocular} ICCV'21 & - & - & - & - & 108.3 & 91.3 & 54.9 \\
    Mesh Graphormer\cite{Lin_2021_ICCV} ICCV'21 & 3D vertices & - & 51.2 & 34.5 & 87.7 & 74.7 & 45.6 \\
    PARE \cite{Kocabas_2021_ICCV} ICCV'21 & Segmentation & - & - & - & 88.6 & 74.5 & 46.5 \\
    THUNDR \cite{zanfir2021thundr} ICCV'21 & 3D markers & - & 55.0 & 39.8 & 88.0 & 74.8 & 51.5 \\
    PyMaf \cite{zhang2021pymaf} ICCV'21 & IUV image & - & 57.7 & 40.5 & 110.1 & 92.8 & 58.9 \\
    ProHMR \cite{Kolotouros_2021_ICCV} ICCV'21 & - & - & - & 41.2 & - & - & 59.8\\
    OCHMR \cite{Khirodkar_2022_CVPR} CVPR'22 & 2D heatmap & - & - & - & 107.1 & 89.7 & 58.3 \\
    3DCrowdNet \cite{Choi_2022_CVPR} CVPR'22 & 3D skeleton & - & - & - & 98.3 & 81.7 & 51.5 \\
    CLIFF \cite{li2022cliff} ECCV'22 & - & - & \textbf{47.1} & 32.7 & 81.2 & 69.0 & 43.0 \\
    FastMETRO \cite{cho2022FastMETRO} ECCV'22 & 3D vertices & - & 52.2 & 33.7 & 84.1 & 73.5 & 44.6 \\
    VisDB \cite{yao2022learning} ECCV'22 & 3D vertices & - & 51.0 & 34.5 & 85.5 & 73.5 & 44.9 \\

    \rowcolor{mygray}
    \textbf{Ours} & Virtual marker & \textbf{58.0} & 47.3 & \textbf{32.0} & \textbf{77.9} & \textbf{67.5} & \textbf{41.3} \\

    \hline 
\end{tabular}}
\caption{Comparison to the state-of-the-arts on H3.6M \cite{h36m_pami} and 3DPW \cite{vonMarcard2018} datasets. $^{\dagger}$ means using temporal cues. The methods are not strictly comparable because they may have different backbones and training datasets. We provide the numbers only to show proof-of-concept results.}
\label{tab:state_of_the_art}
\vspace{-0.4cm}
\end{table*}

\subsection{Datasets and metrics}
\label{subsec:dataset}
\noindent\textbf{H3.6M \cite{h36m_pami}.} We use (S1, S5, S6, S7, S8) for training and (S9, S11) for testing. 
As in \cite{kanazawa2018end, choi2020pose2mesh, lin2021end, Lin_2021_ICCV}, we report MPJPE and PA-MPJPE for poses that are derived from the estimated meshes. We also report Mean Per Vertex Error (MPVE) for the whole mesh.  \\

\noindent\textbf{3DPW \cite{vonMarcard2018}} is collected in natural scenes. 
Following the previous works \cite{lin2021end, Lin_2021_ICCV, Kocabas_2021_ICCV, zanfir2021thundr}, we use the train set of 3DPW to learn the model and evaluate on the test set. The same evaluation metrics as H3.6M are used.  \\

\noindent\textbf{SURREAL \cite{varol2017learning}} is a large-scale synthetic dataset with GT SMPL annotations and has diverse samples in terms of body shapes, backgrounds, \etc. We use its training set to train a model and evaluate the test split following \cite{choi2020pose2mesh}.

\subsection{Implementation Details}
\label{subsec:implementation}
We learn $64$ virtual markers on the H3.6M \cite{h36m_pami} training set. We use the same set of markers for all datasets instead of learning a separate set on each dataset. Following \cite{kanazawa2018end, choi2020pose2mesh, moon2020i2l, zanfir2021thundr, kolotouros2019convolutional, kocabas2020vibe, Lin_2021_ICCV, lin2021end}, we conduct mix-training by using MPI-INF-3DHP \cite{mehta2017monocular}, UP-3D \cite{lassner2017unite}, and COCO \cite{lin2014microsoft} training set for experiments on the H3.6M and 3DPW datasets. 
We adapt a 3D pose estimator \cite{sun2018integral} with HRNet-W48 \cite{sun2019deep} as the image feature backbone for estimating the 3D virtual markers. We set the number of voxels in each dimension to be $64$, \ie $D = H = W = 64$ for 3D heatmaps. Following \cite{kanazawa2018end, kolotouros2019convolutional, moon2020i2l}, we crop every single human region from the input image and resize it to $256 \times 256$.
We use Adam \cite{kingma2015adam} optimizer to train the whole framework for $40$ epochs with a batch size of $32$. The learning rates for the two branches are set to $5 \times 10^{-4}$ and $1 \times 10^{-3}$, respectively, which are decreased by half after the $30^{th}$ epoch.
Please refer to the supplementary for more details.

\begin{table}[t]
\center
\small
\setlength{\tabcolsep}{2pt}
\resizebox{3.4in}{!}{
\begin{tabular}{l l | c c c}
    \hline
    \multirow{2}{*}{Method} & Intermediate & \multirow{2}{*}{MPVE$\downarrow$} & \multirow{2}{*}{MPJPE$\downarrow$} & \multirow{2}{*}{PA-MPJPE$\downarrow$} \\
    & Representation &  &  &  \\
    \hline
    HMR \cite{kanazawa2018end} CVPR'18 & - & 85.1 & 73.6 & 55.4 \\
    BodyNet \cite{varol2018bodynet} ECCV'18 & Skel. + Seg. & 65.8 & - & - \\
    GraphCMR \cite{kolotouros2019convolutional} CVPR'19 & 3D vertices & 103.2 & 87.4 & 63.2  \\
    SPIN \cite{kolotouros2019learning} ICCV'19 & - & 82.3 & 66.7 & 43.7 \\
    DecoMR \cite{zeng20203d} CVPR'20 & IUV image & 68.9 & 52.0 & 43.0 \\
    Pose2Mesh \cite{choi2020pose2mesh} ECCV'20 & 3D skeleton & 68.8 & 56.6 & 39.6 \\
    PC-HMR \cite{luan2021pc} AAAI'21 & 3D skeleton & 59.8 & 51.7 & 37.9  \\
    $^{*}$ DynaBOA \cite{guan2022out} TPAMI'22 & - & 70.7 & 55.2 & 34.0 \\
    \rowcolor{mygray}
    \textbf{Ours} & Virtual marker & \textbf{44.7} & \textbf{36.9} & \textbf{28.9} \\
    
    \hline 
\end{tabular}}
\caption{Comparison to the state-of-the-arts on SURREAL \cite{varol2017learning} dataset. $^{*}$ means training on the test split with 2D supervisions. ``Skel. + Seg.'' means using skeleton and segmentation together.}
\label{tab:state_of_the_art_surreal}
\vspace{-0.4cm}
\end{table}

\subsection{Comparison to the State-of-the-arts}
\label{subsec:sota}

\noindent\textbf{Results on H3.6M.}
Table \ref{tab:state_of_the_art} compares our approach to the state-of-the-art methods on the H3.6M dataset. Our method achieves competitive or superior performance. In particular, it outperforms the methods that use skeletons (Pose2Mesh \cite{choi2020pose2mesh}, DSD-SATN \cite{sun2019human}), body markers (THUNDR) \cite{zanfir2021thundr}, or IUV image \cite{zeng20203d, zhang2021pymaf} as proxy representations, demonstrating the effectiveness of the virtual marker representation. \\

\noindent\textbf{Results on 3DPW.}
We compare our method to the state-of-the-art methods on the 3DPW dataset in Table \ref{tab:state_of_the_art}. Our approach achieves state-of-the-art results among all the methods, validating the advantages of the virtual marker representation over the skeleton representation used in Pose2Mesh \cite{choi2020pose2mesh}, DSD-SATN \cite{sun2019human}, and other representations like IUV image used in PyMAF \cite{zhang2021pymaf}. In particular, our approach outperforms I2L-MeshNet \cite{moon2020i2l}, METRO \cite{lin2021end}, and Mesh Graphormer \cite{Lin_2021_ICCV} by a notable margin, which suggests that 
virtual markers are more suitable and effective representations than detecting all vertices directly as most of them are not discriminative enough to be accurately detected. \\

\noindent\textbf{Results on SURREAL.}
This dataset has more diverse samples in terms of body shapes. The results are shown in Table \ref{tab:state_of_the_art_surreal}. Our approach outperforms the state-of-the-art methods by a notable margin, especially in terms of MPVE. Figure \ref{fig:surreal_shape} shows some challenging cases without cherry-picking. The skeleton representation loses the body shape information so the method \cite{choi2020pose2mesh} can only recover mean shapes. In contrast, our approach generates much more accurate mesh estimation results. 

\begin{table}[t]
\center
\small
\setlength{\tabcolsep}{10pt}
\resizebox{2.8in}{!}{
\begin{tabular}{c | l | c | c }
    \hline 
    \multirow{2}{*}{No.} & Intermediate & \multicolumn{2}{c}{MPVE$\downarrow$} \\

    \cline{3-4}
    & Representation & H3.6M & SURREAL \\
    \hline
    (a) & Skeleton & 64.4 & 53.6 \\ 
    (b) & Rand virtual marker & 63.0 & 50.1 \\
    \rowcolor{mygray}
    (c) & Virtual marker & \textbf{58.0} & \textbf{44.7}\\
    \hline 
\end{tabular}}
\caption{Ablation study of the virtual marker representation for our approach on H3.6M and SURREAL datasets. 
``Skeleton'' means the sparse landmark joint representation is used. 
``Rand virtual marker'' means the virtual markers are randomly selected from all the vertices without learning. (c) is our method, where the learned virtual markers are used. }
\label{tab:ba_effect}
\vspace{-0.4cm}
\end{table}

\subsection{Ablation study}
\label{subsec:ablation}
\noindent\textbf{Virtual marker representation.}
\label{subsubsec:ablation_effect}
We compare our method to two baselines in Table \ref{tab:ba_effect}. First, in baseline (a), we replace the virtual markers of our method with the skeleton representation. The rest are kept the same as ours (c). Our method achieves a much lower MPVE than the baseline (a), demonstrating that the virtual markers help to estimate body shapes more accurately than the skeletons. In baseline (b), we randomly sample $64$ from the $6890$ mesh vertices as virtual markers. We repeat the experiment five times and report the average number. We can see that the result is worse than ours, which is because the randomly selected vertices may not be expressive to reconstruct the other vertices or can not be accurately detected from images as they lack distinguishable visual patterns. The results validate the effectiveness of our learning strategy.

\begin{figure}[t]
	\centering
	\includegraphics[width=3.2in]{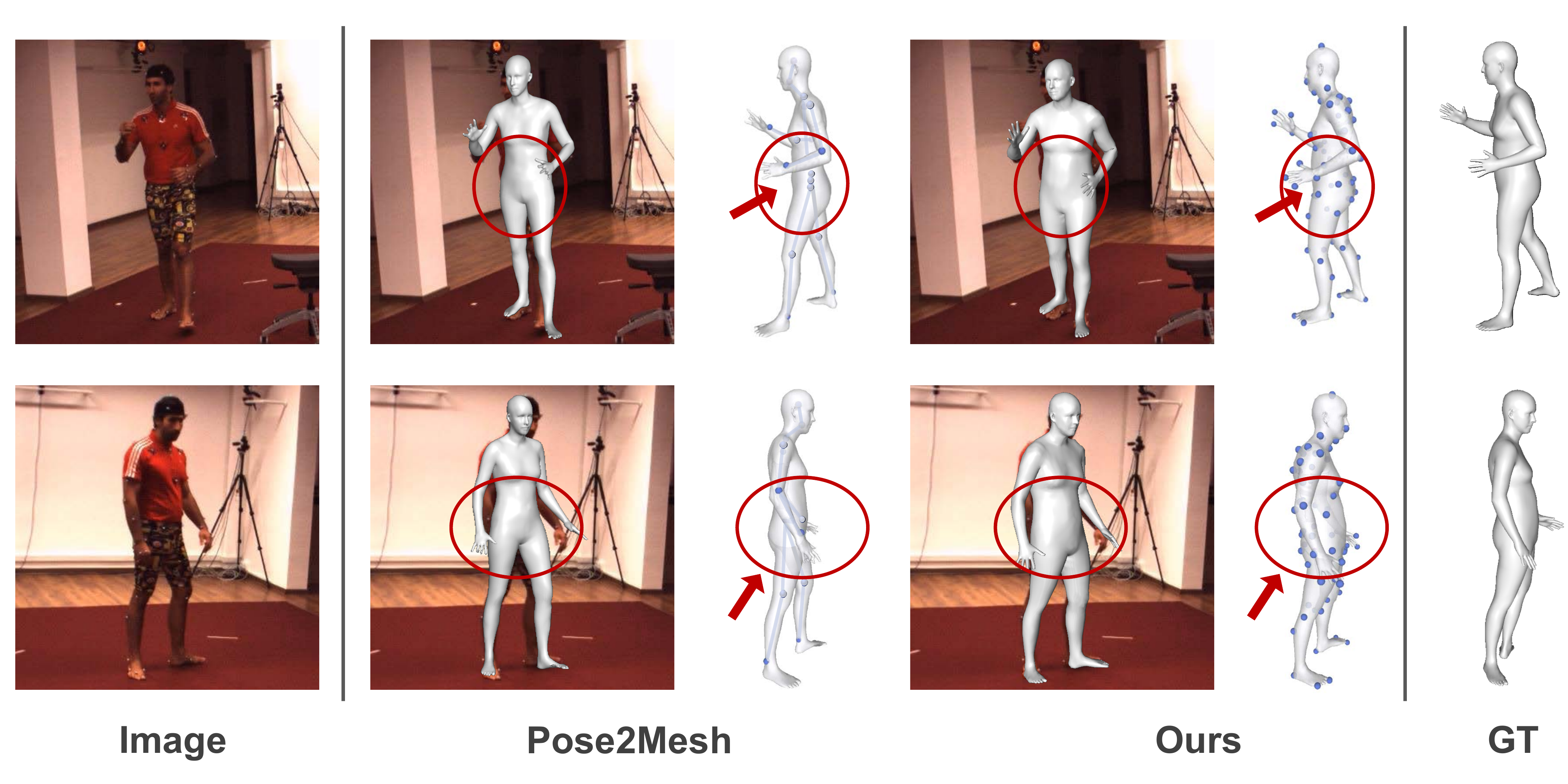}
	\caption{Mesh estimation results of different methods on H3.6M test set. Our method with virtual marker representation gets better shape estimation results than Pose2Mesh which uses skeleton representation. Note the waistline of the body and the thickness of the arm.}
	\label{fig:joint_diff}
\vspace{-0.4cm}
\end{figure}

Figure \ref{fig:surreal_shape} shows some qualitative results on the SURREAL test set. The meshes estimated by the baseline which uses skeleton representation, \ie Pose2Mesh \cite{choi2020pose2mesh}, have inaccurate body shapes. This is reasonable because the skeleton is oversimplified and has very limited capability to recover shapes. Instead, it implicitly learns a mean shape for the whole training dataset. In contrast, the mesh estimated by using virtual markers has much better quality due to its strong representation power and therefore can handle different body shapes elegantly. 
Figure \ref{fig:joint_diff} also shows some qualitative results on the H3.6M test set. For clarity, we draw the intermediate representation (blue balls) in it as well.

\begin{figure}[t]
	\centering
	\includegraphics[width=3.3in]{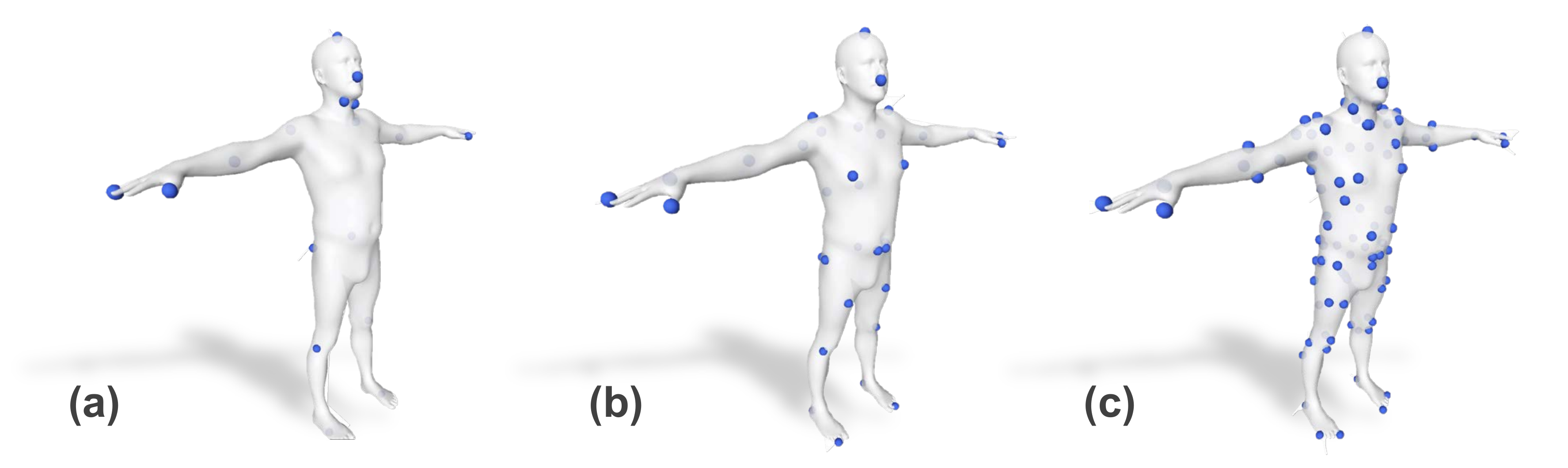}
	\caption{Visualization of the learned virtual markers of different numbers of $K = 16, 32, 96$, from left to right, respectively.
    }
	\label{fig:different_K_vis}
\end{figure}

\vspace{0.5em}
\noindent\textbf{Number of virtual markers.} 
\label{subsubsec:ablation_K}
We evaluate how the number of virtual markers affects estimation quality on H3.6M \cite{h36m_pami} dataset. Figure \ref{fig:different_K_vis} visualizes the learned virtual markers, which are all located on the body surface and close to the extreme points of the mesh. This is expected as mentioned in Section \ref{subsec:body_arche}.  Table \ref{tab:different_K} (GT) shows the mesh reconstruction results when we have GT 3D positions of the virtual markers in objective (\ref{eq:aa}). When we increase the number of virtual markers, both mesh reconstruction error (MPVE) and the regressed landmark joint error (MPJPE) steadily decrease. This is expected because using more virtual markers improves the representation power. However, using more virtual markers cannot guarantee smaller estimation errors when we need to estimate the virtual marker positions from images as in our method. This is because the additional virtual markers may have large estimation errors which affect the mesh estimation result. The results are shown in Table \ref{tab:different_K} (Det). Increasing the number of virtual markers $K$ steadily reduces the MPVE errors when $K$ is smaller than $96$. However, if we keep increasing $K$, the error begins to increase. This is mainly because some of the newly introduced virtual markers are difficult to detect from images and therefore bring errors to mesh estimation.

\begin{table}[t]
\center
\small
\setlength{\tabcolsep}{13pt}
\resizebox{3.0in}{!}{
\begin{tabular}{c | c | c | c | c }
    \hline 
    \multirow{2}{*}{$K$} & \multicolumn{2}{c|}{GT}  & \multicolumn{2}{c}{Det} \\

    \cline{2-3} \cline{4-5}
    & MPVE$\downarrow$ & MPJPE$\downarrow$ & MPVE$\downarrow$ & MPJPE$\downarrow$\\
    \hline
    16 & 46.8 & 39.8 & 58.7 & 47.8 \\ 
    32 & 20.1 & 14.2 & 58.2 & 48.3 \\
    64 & 11.0 & 7.5 & \textbf{58.0} & \textbf{47.3} \\
    96 & \textbf{9.9} & \textbf{5.6} & 59.6 & 48.2\\
    \hline 
\end{tabular}}
\caption{Ablation study of the different number of virtual markers ($K$) on H3.6M \cite{h36m_pami} dataset. (GT) Mesh reconstruction results when GT 3D positions of the virtual markers are used in objective (\ref{eq:aa}). (Det) Mesh estimation results obtained by our proposed framework when we use different numbers of virtual markers ($K$). 
}
\label{tab:different_K}
\end{table}

\begin{figure}[t]
	\centering
	\includegraphics[width=3.4in]{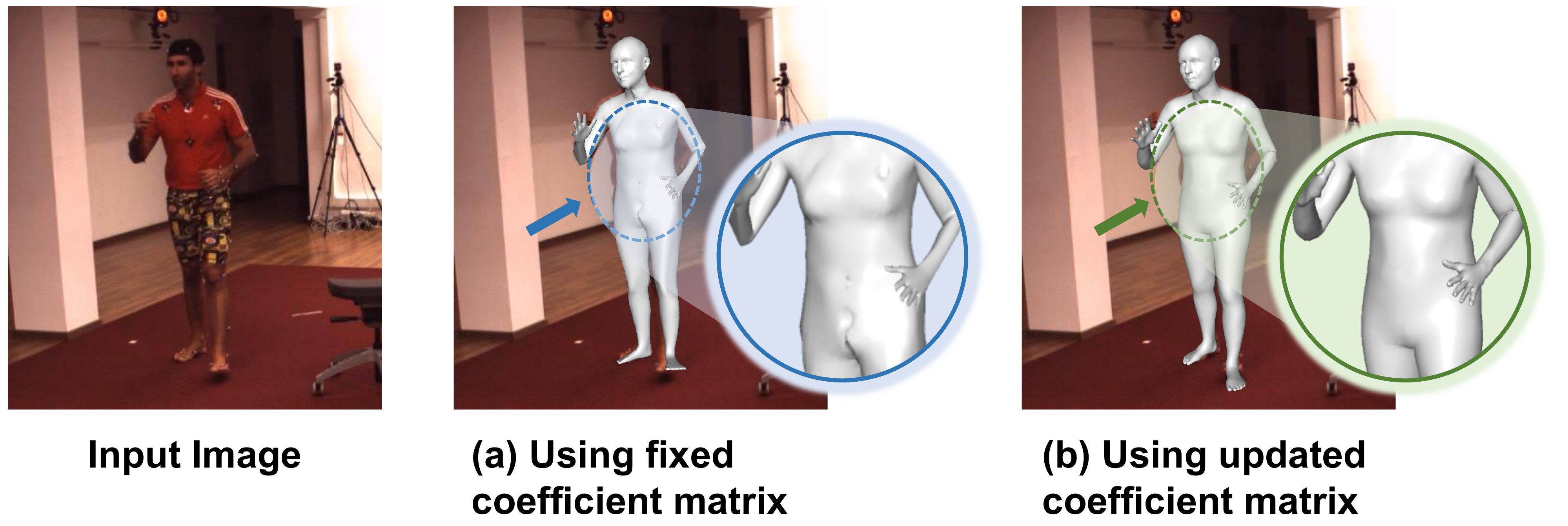}
	\caption{Mesh estimation comparison results when using (a) fixed coefficient matrix $\widetilde{\mathbf{A}}^{sym}$, and (b) updated $\hat{\mathbf{A}}$. 
    Please zoom in to better see the details. 
    }
	\label{fig:blending_nr_diff}
\end{figure}

\begin{figure*}[t]
	\centering
	\includegraphics[width=6.8in]{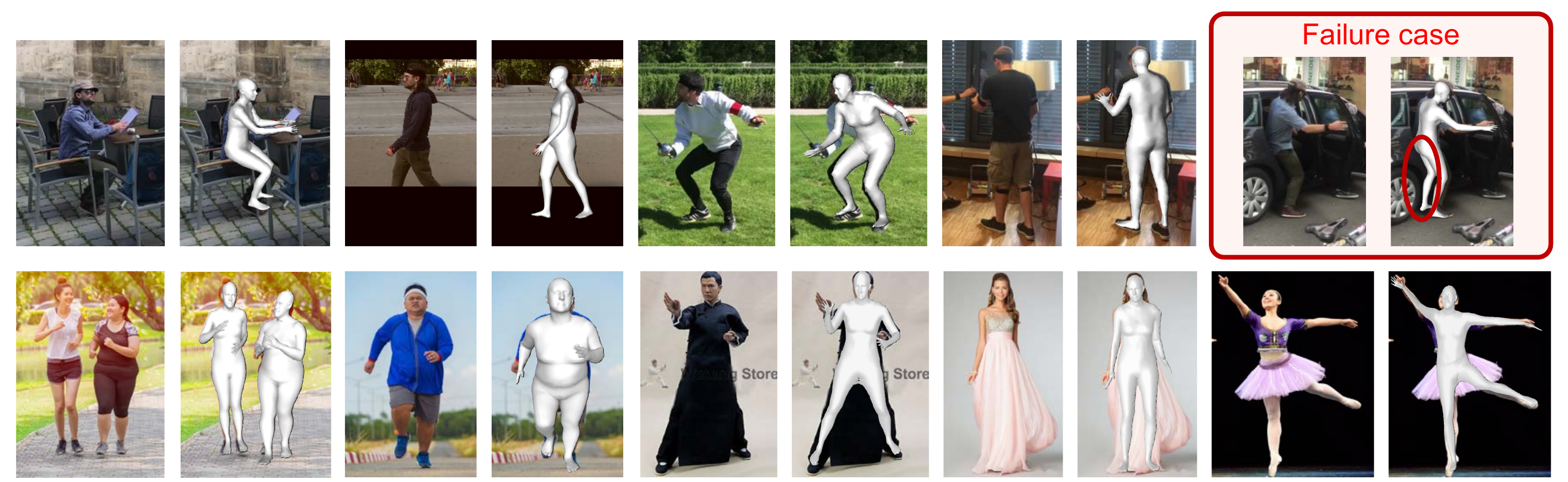}
	\caption{ \textbf{Top:} Meshes estimated by our approach on images from 3DPW test set. The rightmost case in the dashed box shows a typical failure. \textbf{Bottom:} Meshes estimated by our approach on Internet images with challenging cases (extreme shapes or in a long dress).}
	\label{fig:quality_result_w_failure}
\vspace{-0.3cm}
\end{figure*}

\vspace{0.5em}
\noindent\textbf{Coefficient matrix.}
\label{subsubsec:blending_effect}
We compare our method to a baseline which uses the fixed coefficient matrix $\widetilde{\mathbf{A}}^{sym}$. 
We show the quality comparison in Figure \ref{fig:blending_nr_diff}. We can see that the estimated mesh by a fixed coefficient matrix (a) has mostly correct pose and shape but there are also some artifacts on the mesh while using the updated coefficient matrix (b) can get better mesh estimation results. 
As shown in Table \ref{tab:quan_bm_nr_effect}, using a fixed coefficient matrix gets larger MPVE and MPJPE errors than using the updated coefficient matrix. This is caused by the estimation errors of virtual markers when occlusion happens, which is inevitable since the virtual markers on the back will be self-occluded by the front body. As a result, inaccurate marker positions would bring large errors to the final mesh estimates if we directly use the fixed matrix.

\begin{table}[]
\center
\small
\setlength{\tabcolsep}{4pt}
\renewcommand\arraystretch{1.35}
\resizebox{3.0in}{!}{
\begin{tabular}{l l | c c |c c}
    \hline 
    No. & Method & Fixed $\widetilde{\mathbf{A}}^{sym}$ & Updated $\hat{\mathbf{A}}$ & MPVE$\downarrow$ & MPJPE$\downarrow$ \\
    \hline
    (a) & Ours (fixed) & \cmark & \xmark  & 64.7 & 51.6 \\
    \rowcolor{mygray}
    (b) & Ours  & \xmark & \cmark  & \textbf{58.0} & \textbf{47.3} \\
    \hline 
\end{tabular}}
\caption{Ablation study of the coefficient matrix for our approach on H3.6M dataset. ``fixed'' means using the fixed coefficient matrix $\widetilde{\mathbf{A}}^{sym}$ to reconstruct the mesh. 
}
\label{tab:quan_bm_nr_effect}
\end{table}

\subsection{Qualitative Results}

\label{subsec:quality}
\begin{figure}[t]
	\centering
	\includegraphics[width=3.0in]{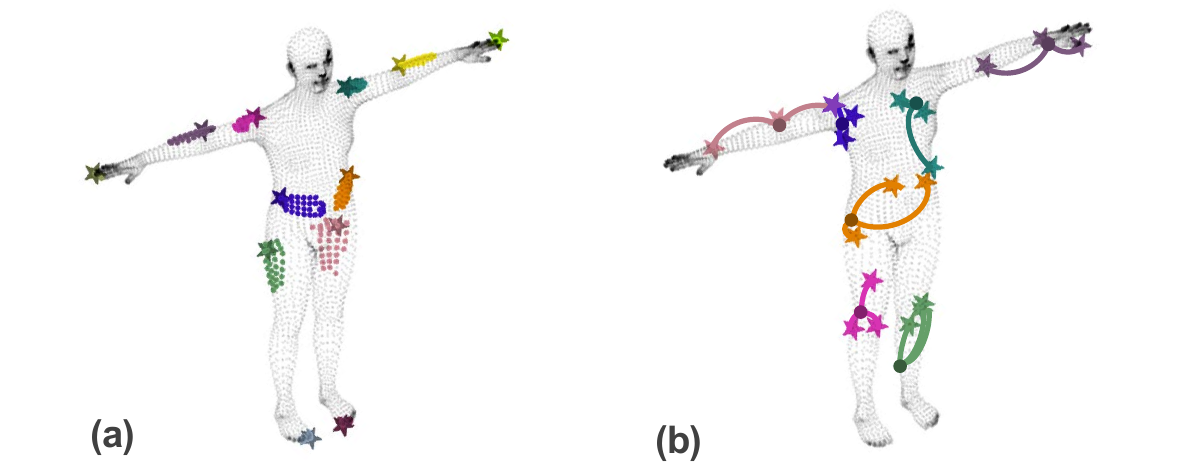}
	\caption{(a) For each virtual marker (represented by a star), we highlight the top 30 most affected vertices (represented by a colored dot) based on average coefficient matrix $\mathbf{\hat{A}}$. (b) For each vertex (dot), we highlight the top 3 virtual markers (star) that contribute the most. We can see that the dependency has a strong locality which improves the robustness when some virtual markers cannot be accurately detected. }
	\label{fig:locality}
\vspace{-0.3cm}
\end{figure}

Figure \ref{fig:quality_result_w_failure} (top) presents some meshes estimated by our approach on natural images from the 3DPW test set. The rightmost case shows a typical failure where our method has a wrong pose estimate of the left leg due to heavy occlusion. We can see that the failure is constrained to the local region and the rest of the body still gets accurate estimates. We further analyze how inaccurate virtual markers would affect the mesh estimation, \ie when part of human body is occluded or truncated. According to the finally learned coefficient matrix $\mathbf{\hat{A}}$ of our model, we highlight the relationship weights among virtual markers and all vertices in Figure \ref{fig:locality}. We can see that our model actually learns \emph{local and sparse} dependency between each vertex and the virtual markers, \eg for each vertex, the virtual markers that contribute the most are in a near range as shown in Figure \ref{fig:locality} (b). Therefore, in inference, if a virtual marker has inaccurate position estimation due to occlusion or truncation, the dependent vertices may have inaccurate estimates, while the rest will be barely affected. Figure \ref{fig:body_arche} (right) shows more examples where occlusion or truncation occurs, and our method can still get accurate or reasonable estimates robustly. Note that when truncation occurs, our method still guesses the positions of the truncated virtual markers.

Figure \ref{fig:quality_result_w_failure} (bottom) shows our estimated meshes on challenging cases, which indicates the strong generalization ability of our model on diverse postures and actions in natural scenes. Please refer to the supplementary for more quality results. Note that since the datasets do not provide supervision of head orientation, face expression, hands, or feet, the estimates of these parts are just in canonical poses inevitably. Apart from that, most errors are due to inaccurate 3D virtual marker estimation which may be addressed using more powerful estimators or more diverse training datasets in the future.

\section{Conclusion}
\label{sec:conclusion}
In this paper, we present a novel intermediate representation \emph{Virtual Marker}, which is more expressive than the prevailing skeleton representation and more accessible than physical markers. It can reconstruct 3D meshes more accurately and efficiently, especially in handling diverse body shapes. Besides, the coefficient matrix in the virtual marker representation encodes spatial relationships among mesh vertices which allows the method to implicitly explore structure priors of human body. It achieves better mesh estimation results than the state-of-the-art methods and shows advanced generalization potential in spite of its simplicity.

\section*{Acknowledgement}  
This work was supported by MOST-2022ZD0114900 and NSFC-62061136001.

\clearpage

{\small
\bibliographystyle{ieee_fullname}
\bibliography{egbib}

\begin{thebibliography}{10}\itemsep=-1pt

\bibitem{arnab2019exploiting}
Anurag Arnab, Carl Doersch, and Andrew Zisserman.
\newblock Exploiting temporal context for 3d human pose estimation in the wild.
\newblock In {\em CVPR}, pages 3395--3404, 2019.

\bibitem{bogo2016keep}
Federica Bogo, Angjoo Kanazawa, Christoph Lassner, Peter Gehler, Javier Romero,
  and Michael~J Black.
\newblock Keep it smpl: Automatic estimation of 3d human pose and shape from a
  single image.
\newblock In {\em ECCV}, pages 561--578, 2016.

\bibitem{boulic1997integration}
Ronan Boulic, Pascal B{\'e}cheiraz, Luc Emering, and Daniel Thalmann.
\newblock Integration of motion control techniques for virtual human and avatar
  real-time animation.
\newblock In {\em Proceedings of the ACM symposium on Virtual reality software
  and technology}, pages 111--118, 1997.

\bibitem{chen2014fast}
Yuansi Chen, Julien Mairal, and Zaid Harchaoui.
\newblock Fast and robust archetypal analysis for representation learning.
\newblock In {\em CVPR}, pages 1478--1485, 2014.

\bibitem{cho2022FastMETRO}
Junhyeong Cho, Kim Youwang, and Tae-Hyun Oh.
\newblock Cross-attention of disentangled modalities for 3d human mesh recovery
  with transformers.
\newblock In {\em ECCV}, 2022.

\bibitem{choi2021beyond}
Hongsuk Choi, Gyeongsik Moon, Ju~Yong Chang, and Kyoung~Mu Lee.
\newblock Beyond static features for temporally consistent 3d human pose and
  shape from a video.
\newblock In {\em CVPR}, pages 1964--1973, 2021.

\bibitem{choi2020pose2mesh}
Hongsuk Choi, Gyeongsik Moon, and Kyoung~Mu Lee.
\newblock Pose2mesh: Graph convolutional network for 3d human pose and mesh
  recovery from a 2d human pose.
\newblock In {\em ECCV}, pages 769--787, 2020.

\bibitem{Choi_2022_CVPR}
Hongsuk Choi, Gyeongsik Moon, JoonKyu Park, and Kyoung~Mu Lee.
\newblock Learning to estimate robust 3d human mesh from in-the-wild crowded
  scenes.
\newblock In {\em CVPR}, pages 1475--1484, June 2022.

\bibitem{choutas2020monocular}
Vasileios Choutas, Georgios Pavlakos, Timo Bolkart, Dimitrios Tzionas, and
  Michael~J Black.
\newblock Monocular expressive body regression through body-driven attention.
\newblock In {\em ECCV}, pages 20--40, 2020.

\bibitem{ci2022gfpose}
Hai Ci, Mingdong Wu, Wentao Zhu, Xiaoxuan Ma, Hao Dong, Fangwei Zhong, and
  Yizhou Wang.
\newblock Gfpose: Learning 3d human pose prior with gradient fields.
\newblock {\em arXiv preprint arXiv:2212.08641}, 2022.

\bibitem{corona2022learned}
Enric Corona, Gerard Pons-Moll, Guillem Aleny{\`a}, and Francesc Moreno-Noguer.
\newblock Learned vertex descent: a new direction for 3d human model fitting.
\newblock In {\em ECCV}, pages 146--165. Springer, 2022.

\bibitem{cutler1994archetypal}
Adele Cutler and Leo Breiman.
\newblock Archetypal analysis.
\newblock {\em Technometrics}, 36(4):338--347, 1994.

\bibitem{gower1975generalized}
John~C Gower.
\newblock Generalized procrustes analysis.
\newblock {\em Psychometrika}, 40(1):33--51, 1975.

\bibitem{guan2022out}
Shanyan Guan, Jingwei Xu, Michelle~Z He, Yunbo Wang, Bingbing Ni, and Xiaokang
  Yang.
\newblock Out-of-domain human mesh reconstruction via dynamic bilevel online
  adaptation.
\newblock {\em IEEE TPAMI}, 2022.

\bibitem{huang2017towards}
Yinghao Huang, Federica Bogo, Christoph Lassner, Angjoo Kanazawa, Peter~V
  Gehler, Javier Romero, Ijaz Akhter, and Michael~J Black.
\newblock Towards accurate marker-less human shape and pose estimation over
  time.
\newblock In {\em 3DV}, pages 421--430, 2017.

\bibitem{h36m_pami}
Catalin Ionescu, Dragos Papava, Vlad Olaru, and Cristian Sminchisescu.
\newblock Human3. 6m: Large scale datasets and predictive methods for 3d human
  sensing in natural environments.
\newblock {\em IEEE TPAMI}, 36(7):1325--1339, 2013.

\bibitem{iskakov2019learnable}
Karim Iskakov, Egor Burkov, Victor Lempitsky, and Yury Malkov.
\newblock Learnable triangulation of human pose.
\newblock In {\em ICCV}, pages 7718--7727, 2019.

\bibitem{jolliffe1986principal}
Ian~T Jolliffe.
\newblock Principal components in regression analysis.
\newblock In {\em Principal component analysis}, pages 129--155. Springer,
  1986.

\bibitem{kanazawa2018end}
Angjoo Kanazawa, Michael~J Black, David~W Jacobs, and Jitendra Malik.
\newblock End-to-end recovery of human shape and pose.
\newblock In {\em CVPR}, pages 7122--7131, 2018.

\bibitem{kanazawa2019learning}
Angjoo Kanazawa, Jason~Y Zhang, Panna Felsen, and Jitendra Malik.
\newblock Learning 3d human dynamics from video.
\newblock In {\em CVPR}, pages 5614--5623, 2019.

\bibitem{Khirodkar_2022_CVPR}
Rawal Khirodkar, Shashank Tripathi, and Kris Kitani.
\newblock Occluded human mesh recovery.
\newblock In {\em CVPR}, pages 1715--1725, June 2022.

\bibitem{kingma2015adam}
Diederik~P Kingma and Jimmy Ba.
\newblock Adam: A method for stochastic optimization.
\newblock In {\em ICLR}, 2015.

\bibitem{kocabas2020vibe}
Muhammed Kocabas, Nikos Athanasiou, and Michael~J Black.
\newblock Vibe: Video inference for human body pose and shape estimation.
\newblock In {\em CVPR}, pages 5253--5263, 2020.

\bibitem{Kocabas_2021_ICCV}
Muhammed Kocabas, Chun-Hao~P. Huang, Otmar Hilliges, and Michael~J. Black.
\newblock Pare: Part attention regressor for 3d human body estimation.
\newblock In {\em ICCV}, pages 11127--11137, October 2021.

\bibitem{kolotouros2019learning}
Nikos Kolotouros, Georgios Pavlakos, Michael~J Black, and Kostas Daniilidis.
\newblock Learning to reconstruct 3d human pose and shape via model-fitting in
  the loop.
\newblock In {\em ICCV}, pages 2252--2261, 2019.

\bibitem{kolotouros2019convolutional}
Nikos Kolotouros, Georgios Pavlakos, and Kostas Daniilidis.
\newblock Convolutional mesh regression for single-image human shape
  reconstruction.
\newblock In {\em CVPR}, pages 4501--4510, 2019.

\bibitem{Kolotouros_2021_ICCV}
Nikos Kolotouros, Georgios Pavlakos, Dinesh Jayaraman, and Kostas Daniilidis.
\newblock Probabilistic modeling for human mesh recovery.
\newblock In {\em ICCV}, pages 11605--11614, October 2021.

\bibitem{lassner2017unite}
Christoph Lassner, Javier Romero, Martin Kiefel, Federica Bogo, Michael~J
  Black, and Peter~V Gehler.
\newblock Unite the people: Closing the loop between 3d and 2d human
  representations.
\newblock In {\em CVPR}, pages 6050--6059, 2017.

\bibitem{li2021hybrik}
Jiefeng Li, Chao Xu, Zhicun Chen, Siyuan Bian, Lixin Yang, and Cewu Lu.
\newblock Hybrik: A hybrid analytical-neural inverse kinematics solution for 3d
  human pose and shape estimation.
\newblock In {\em CVPR}, pages 3383--3393, 2021.

\bibitem{li2020pastanet}
Yong-Lu Li, Liang Xu, Xinpeng Liu, Xijie Huang, Yue Xu, Shiyi Wang, Hao-Shu
  Fang, Ze Ma, Mingyang Chen, and Cewu Lu.
\newblock Pastanet: Toward human activity knowledge engine.
\newblock In {\em CVPR}, pages 382--391, 2020.

\bibitem{li2022cliff}
Zhihao Li, Jianzhuang Liu, Zhensong Zhang, Songcen Xu, and Youliang Yan.
\newblock Cliff: Carrying location information in full frames into human pose
  and shape estimation.
\newblock In {\em ECCV}, 2022.

\bibitem{lin2021end}
Kevin Lin, Lijuan Wang, and Zicheng Liu.
\newblock End-to-end human pose and mesh reconstruction with transformers.
\newblock In {\em CVPR}, pages 1954--1963, 2021.

\bibitem{Lin_2021_ICCV}
Kevin Lin, Lijuan Wang, and Zicheng Liu.
\newblock Mesh graphormer.
\newblock In {\em ICCV}, pages 12939--12948, 2021.

\bibitem{lin2014microsoft}
Tsung-Yi Lin, Michael Maire, Serge Belongie, James Hays, Pietro Perona, Deva
  Ramanan, Piotr Doll{\'a}r, and C~Lawrence Zitnick.
\newblock Microsoft coco: Common objects in context.
\newblock In {\em ECCV}, pages 740--755, 2014.

\bibitem{loper2014mosh}
Matthew Loper, Naureen Mahmood, and Michael~J Black.
\newblock Mosh: Motion and shape capture from sparse markers.
\newblock {\em TOG}, 33(6):1--13, 2014.

\bibitem{loper2015smpl}
Matthew Loper, Naureen Mahmood, Javier Romero, Gerard Pons-Moll, and Michael~J
  Black.
\newblock Smpl: A skinned multi-person linear model.
\newblock {\em TOG}, 34(6):1--16, 2015.

\bibitem{luan2021pc}
Tianyu Luan, Yali Wang, Junhao Zhang, Zhe Wang, Zhipeng Zhou, and Yu Qiao.
\newblock Pc-hmr: Pose calibration for 3d human mesh recovery from 2d
  images/videos.
\newblock In {\em AAAI}, pages 2269--2276, 2021.

\bibitem{mehta2017monocular}
Dushyant Mehta, Helge Rhodin, Dan Casas, Pascal Fua, Oleksandr Sotnychenko,
  Weipeng Xu, and Christian Theobalt.
\newblock Monocular 3d human pose estimation in the wild using improved cnn
  supervision.
\newblock In {\em 3DV}, pages 506--516, 2017.

\bibitem{moon2020i2l}
Gyeongsik Moon and Kyoung~Mu Lee.
\newblock I2l-meshnet: Image-to-lixel prediction network for accurate 3d human
  pose and mesh estimation from a single rgb image.
\newblock In {\em ECCV}, pages 752--768, 2020.

\bibitem{omran2018neural}
Mohamed Omran, Christoph Lassner, Gerard Pons-Moll, Peter Gehler, and Bernt
  Schiele.
\newblock Neural body fitting: Unifying deep learning and model based human
  pose and shape estimation.
\newblock In {\em 3DV}, pages 484--494. IEEE, 2018.

\bibitem{pavlakos2019expressive}
Georgios Pavlakos, Vasileios Choutas, Nima Ghorbani, Timo Bolkart, Ahmed~AA
  Osman, Dimitrios Tzionas, and Michael~J Black.
\newblock Expressive body capture: 3d hands, face, and body from a single
  image.
\newblock In {\em CVPR}, pages 10975--10985, 2019.

\bibitem{pavlakos2018learning}
Georgios Pavlakos, Luyang Zhu, Xiaowei Zhou, and Kostas Daniilidis.
\newblock Learning to estimate 3d human pose and shape from a single color
  image.
\newblock In {\em CVPR}, pages 459--468, 2018.

\bibitem{presti20163d}
Liliana~Lo Presti and Marco La~Cascia.
\newblock 3d skeleton-based human action classification: A survey.
\newblock {\em Pattern Recognition}, 53:130--147, 2016.

\bibitem{qiu2019cross}
Haibo Qiu, Chunyu Wang, Jingdong Wang, Naiyan Wang, and Wenjun Zeng.
\newblock Cross view fusion for 3d human pose estimation.
\newblock In {\em ICCV}, pages 4342--4351, 2019.

\bibitem{su2022virtualpose}
Jiajun Su, Chunyu Wang, Xiaoxuan Ma, Wenjun Zeng, and Yizhou Wang.
\newblock Virtualpose: Learning generalizable 3d human pose models from virtual
  data.
\newblock In {\em ECCV}, pages 55--71. Springer, 2022.

\bibitem{sun2019deep}
Ke Sun, Bin Xiao, Dong Liu, and Jingdong Wang.
\newblock Deep high-resolution representation learning for human pose
  estimation.
\newblock In {\em CVPR}, pages 5693--5703, 2019.

\bibitem{sun2018integral}
Xiao Sun, Bin Xiao, Fangyin Wei, Shuang Liang, and Yichen Wei.
\newblock Integral human pose regression.
\newblock In {\em ECCV}, pages 529--545, 2018.

\bibitem{sun2021monocular}
Yu Sun, Qian Bao, Wu Liu, Yili Fu, Michael~J Black, and Tao Mei.
\newblock Monocular, one-stage, regression of multiple 3d people.
\newblock In {\em ICCV}, pages 11179--11188, 2021.

\bibitem{sun2019human}
Yu Sun, Yun Ye, Wu Liu, Wenpeng Gao, Yili Fu, and Tao Mei.
\newblock Human mesh recovery from monocular images via a skeleton-disentangled
  representation.
\newblock In {\em ICCV}, pages 5349--5358, 2019.

\bibitem{tu2020voxelpose}
Hanyue Tu, Chunyu Wang, and Wenjun Zeng.
\newblock Voxelpose: Towards multi-camera 3d human pose estimation in wild
  environment.
\newblock In {\em ECCV}, pages 197--212. Springer, 2020.

\bibitem{NIPS2017_ab452534}
Hsiao-Yu Tung, Hsiao-Wei Tung, Ersin Yumer, and Katerina Fragkiadaki.
\newblock Self-supervised learning of motion capture.
\newblock In {\em NIPS}, volume~30, 2017.

\bibitem{varol2018bodynet}
Gul Varol, Duygu Ceylan, Bryan Russell, Jimei Yang, Ersin Yumer, Ivan Laptev,
  and Cordelia Schmid.
\newblock Bodynet: Volumetric inference of 3d human body shapes.
\newblock In {\em ECCV}, pages 20--36, 2018.

\bibitem{varol2017learning}
Gul Varol, Javier Romero, Xavier Martin, Naureen Mahmood, Michael~J Black, Ivan
  Laptev, and Cordelia Schmid.
\newblock Learning from synthetic humans.
\newblock In {\em CVPR}, pages 109--117, 2017.

\bibitem{vonMarcard2018}
Timo von Marcard, Roberto Henschel, Michael~J Black, Bodo Rosenhahn, and Gerard
  Pons-Moll.
\newblock Recovering accurate 3d human pose in the wild using imus and a moving
  camera.
\newblock In {\em ECCV}, pages 601--617, 2018.

\bibitem{wan2021encoder}
Ziniu Wan, Zhengjia Li, Maoqing Tian, Jianbo Liu, Shuai Yi, and Hongsheng Li.
\newblock Encoder-decoder with multi-level attention for 3d human shape and
  pose estimation.
\newblock In {\em ICCV}, pages 13033--13042, 2021.

\bibitem{wang2018pixel2mesh}
Nanyang Wang, Yinda Zhang, Zhuwen Li, Yanwei Fu, Wei Liu, and Yu-Gang Jiang.
\newblock Pixel2mesh: Generating 3d mesh models from single rgb images.
\newblock In {\em ECCV}, pages 52--67, 2018.

\bibitem{xu2019denserac}
Yuanlu Xu, Song-Chun Zhu, and Tony Tung.
\newblock Denserac: Joint 3d pose and shape estimation by dense
  render-and-compare.
\newblock In {\em ICCV}, pages 7760--7770, 2019.

\bibitem{yao2022learning}
Chun-Han Yao, Jimei Yang, Duygu Ceylan, Yi Zhou, Yang Zhou, and Ming-Hsuan
  Yang.
\newblock Learning visibility for robust dense human body estimation.
\newblock In {\em ECCV}, 2022.

\bibitem{ye2022faster}
Hang Ye, Wentao Zhu, Chunyu Wang, Rujie Wu, and Yizhou Wang.
\newblock Faster voxelpose: Real-time 3d human pose estimation by orthographic
  projection.
\newblock In {\em ECCV}, pages 142--159. Springer, 2022.

\bibitem{zanfir2018monocular}
Andrei Zanfir, Elisabeta Marinoiu, and Cristian Sminchisescu.
\newblock Monocular 3d pose and shape estimation of multiple people in natural
  scenes-the importance of multiple scene constraints.
\newblock In {\em CVPR}, pages 2148--2157, 2018.

\bibitem{zanfir2021thundr}
Mihai Zanfir, Andrei Zanfir, Eduard~Gabriel Bazavan, William~T Freeman, Rahul
  Sukthankar, and Cristian Sminchisescu.
\newblock Thundr: Transformer-based 3d human reconstruction with markers.
\newblock In {\em ICCV}, pages 12971--12980, 2021.

\bibitem{zeng20203d}
Wang Zeng, Wanli Ouyang, Ping Luo, Wentao Liu, and Xiaogang Wang.
\newblock 3d human mesh regression with dense correspondence.
\newblock In {\em CVPR}, pages 7054--7063, 2020.

\bibitem{zhang2020densepose2smpl}
Hongwen Zhang, Jie Cao, Guo Lu, Wanli Ouyang, and Zhenan Sun.
\newblock Learning 3d human shape and pose from dense body parts.
\newblock {\em IEEE TPAMI}, 44(5):2610--2627, 2022.

\bibitem{zhang2021pymaf}
Hongwen Zhang, Yating Tian, Xinchi Zhou, Wanli Ouyang, Yebin Liu, Limin Wang,
  and Zhenan Sun.
\newblock Pymaf: 3d human pose and shape regression with pyramidal mesh
  alignment feedback loop.
\newblock In {\em ICCV}, pages 11446--11456, 2021.

\bibitem{zhang2022voxeltrack}
Yifu Zhang, Chunyu Wang, Xinggang Wang, Wenyu Liu, and Wenjun Zeng.
\newblock Voxeltrack: Multi-person 3d human pose estimation and tracking in the
  wild.
\newblock {\em IEEE TPAMI}, 45(2):2613--2626, 2022.

\end{thebibliography}
}

\newpage
\section*{Appendix}

We elaborate on the post-processing implementation of the virtual markers and provide additional experimental details and results. At last, we discuss data from human subjects and the potential societal impact.

\subsection*{A. \quad Post-processing on Virtual Markers}
\label{sec:post}

As described in Section \ref{subsec:body_arche}, considering the left-right symmetric human body structure, we slightly adjust the learned virtual markers $\mathbf{Z}$ to be symmetric. In fact, after the first step that updates each $\mathbf{z}_i$ by its nearest vertex to get $\widetilde{\mathbf{Z}} \in \mathbb{R}^{3 \times K}$. $\widetilde{\mathbf{Z}}$ are almost symmetric with few exceptions. To get the final symmetric virtual markers $\widetilde{\mathbf{Z}}^{sym} \in \mathbb{R}^{3 \times K}$, for each virtual marker located in the left body part, we take its symmetric vertex in the right body to be its symmetric counterpart. 

\begin{figure}[h]
    \centering
    \includegraphics[width=1.8in]{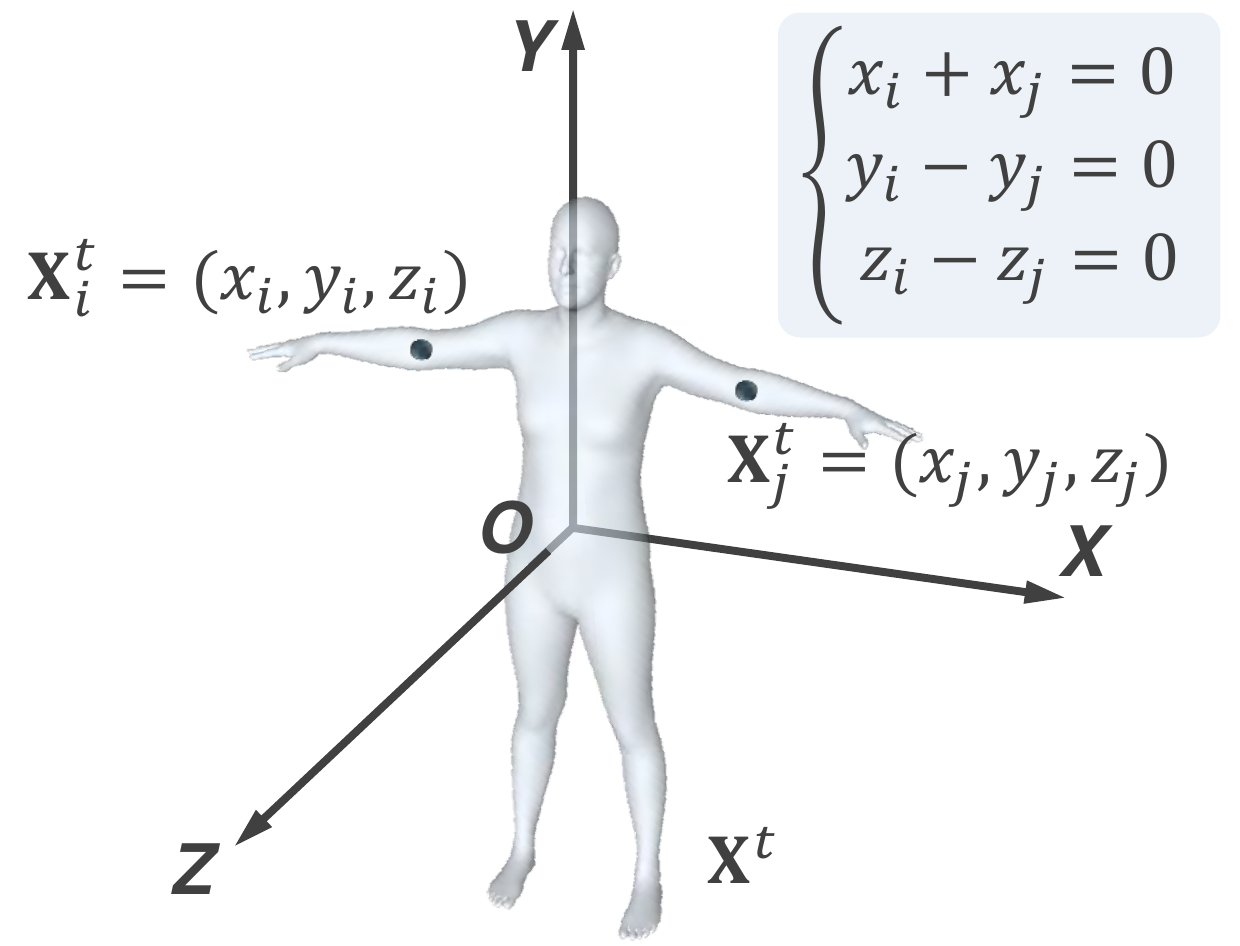}
    \caption{Illustration of the human mesh template $\mathbf{X}^{t}$ at the 3D coordinate system and a symmetric vertex pair $(\mathbf{X}^t_{i}, \mathbf{X}^t_{j})$.}
    \label{fig:supp_mesh}
\end{figure}

Since the human mesh (\ie SMPL \cite{loper2015smpl}) itself is not strictly symmetric, we clarify the \emph{symmetric vertex pair} (\eg left elbow and right elbow) on a human mesh template $\mathbf{X}^t \in \mathbb{R}^{3 \times M}$ in Figure \ref{fig:supp_mesh}. We place $\mathbf{X}^{t}$ at the origin of the 3D coordinate system. Formally, we define the cost of matching $i^{th}$ vertex to $j^{th}$ vertex to be $\bm{C}_{i, j} = \left|x_i + x_j\right| + \left|y_i - y_j\right| + \left|z_i - z_j\right|$. A symmetric vertex pair $(\mathbf{X}^t_{i}, \mathbf{X}^t_{j})$ is defined to have the minimal cost $\bm{C}_{i, j}$. In this way, for each virtual marker in the left body, we take its symmetric vertex counterpart to be its symmetric virtual marker and finally get $\widetilde{\mathbf{Z}}^{sym}$.

\subsection*{B. \quad Experiments}
\label{sec:supp_experiments}

In this section, we first add detailed descriptions for datasets and then provide more experimental results of our approach.

\subsubsection*{B.1 \quad Datasets}

\paragraph{H3.6M \cite{h36m_pami}.} Following previous works \cite{kanazawa2018end, kolotouros2019learning, kolotouros2019convolutional, pavlakos2018learning}, we use the SMPL parameters generated from MoSh \cite{loper2014mosh}, which are fitted to the 3D physical marker locations, to get the GT 3D mesh supervision. Following standard practice \cite{kanazawa2018end}, we evaluate the quality of 3D pose of $14$ joints derived from the estimated mesh, \ie $\hat{\mathbf{M}}\mathcal{J}$. We report Mean Per Joint Position Error (MPJPE) and PA-MPJPE in millimeters (mm). The latter uses Procrustes algorithm \cite{gower1975generalized} to align the estimates to GT poses before computing MPJPE. To evaluate mesh estimation results, we also report Mean Per Vertex Error (MPVE) which can be interpreted as MPJPE computed over the whole mesh.

\paragraph{3DPW \cite{vonMarcard2018}.} The 3D GT SMPL parameters are obtained by using the data from IMUs when collected. Following the previous works \cite{lin2021end, Lin_2021_ICCV, Kocabas_2021_ICCV, zanfir2021thundr}, we use the train set of 3DPW to learn the model and evaluate on the test set.

\paragraph{MPI-INF-3DHP \cite{mehta2017monocular}} is a 3D pose dataset with 3D GT pose annotations. Since this dataset does not provide 3D mesh annotations, following \cite{kanazawa2018end, kolotouros2019learning}, we only enforce supervision on the 3D skeletons (Eq. (\ref{eq:loss_reg})) in mesh losses.

\paragraph{UP-3D \cite{lassner2017unite}} is a wild 2D pose dataset with natural images. The 3D poses and meshes are obtained by SMPLify \cite{bogo2016keep}. Due to the lack of GT 3D poses, the fitted meshes are not accurate. Therefore we only use the 2D annotations to train the 3D virtual marker estimation network as in \cite{sun2018integral}.  

\paragraph{COCO \cite{lin2014microsoft}} is a large wild 2D pose dataset with natural images. Previous work \cite{moon2020i2l} used SMPLify-X \cite{pavlakos2019expressive} to obtain pseudo SMPL mesh annotations but they are not accurate. However, we find that if we project the 3D mesh to 2D image, the resulting 2D mesh vertices align well with the image. So we leverage the 2D annotations to train the virtual marker estimation network as in \cite{sun2018integral}.

\paragraph{SURREAL \cite{varol2017learning}} is a large-scale synthetic dataset containing 6 million frames of synthetic humans. The images are photo-realistic renderings of people under large variations in shape, texture, viewpoint, and body pose. To ensure realism, the synthetic bodies are created using the SMPL body model, whose parameters are fit by the MoSh \cite{loper2014mosh} given raw 3D physical marker data. All the images have a resolution of $320 \times 240$. We use the same training split to train the model and evaluate the test split following \cite{choi2020pose2mesh}.

\subsubsection*{B.2 \quad Implementation Details and Computation Resource}
Following common practice \cite{kanazawa2018end, choi2020pose2mesh, moon2020i2l, zanfir2021thundr, kolotouros2019convolutional, kocabas2020vibe, Lin_2021_ICCV, lin2021end}, we conduct mix-training by using the above 2D and 3D datasets for experiments on the H3.6M and 3DPW datasets. 
To leverage the 3D pose estimation dataset, \ie MPI-INF-3DHP \cite{mehta2017monocular}, we extend the $64$ virtual markers with the $17$ landmark joints (\ie skeleton) from the MPI-INF-3DHP dataset. 
For experiments on the SURREAL dataset, we use its training set alone as in \cite{choi2020pose2mesh, luan2021pc}. We implement the proposed method with PyTorch. All the experiments are conducted on a Linux machine with 4 NVIDIA 16GB V100 GPUs. The whole network is trained for 40 epochs with batch size 32 using Adam \cite{kingma2015adam} optimizer.

We evaluate the model complexity in terms of FLOPs (G) and the number of model parameters in Table \ref{tab:computation}. Compared to the most recent state-of-the-art methods that directly regress \textit{all mesh vertices}, such as I2L-MeshNet \cite{moon2020i2l}, METRO \cite{lin2021end}, and Mesh Graphormer \cite{Lin_2021_ICCV}, our approach with virtual marker representation reduces the computation overhead by a large margin while getting better estimation quality. The last column shows the MPVE errors on 3DPW test set for performance reference.

\begin{table}[t]
\center
\small
\setlength{\tabcolsep}{4pt}
\resizebox{3.2in}{!}{
\begin{tabular}{l | c  c | c}
    \hline 
    Methods & FLOPs (G) $\downarrow$ & Params (M) & MPVE$\downarrow$\\

    \hline
    I2L-MeshNet \cite{moon2020i2l} ECCV'20 & 28.7 & 141.2 & 110.1\\ 
    METRO \cite{lin2021end} CVPR'21 & 153.0 & 397.5 & 88.2\\ 
    Mesh Graphormer \cite{Lin_2021_ICCV} ICCV'21 & 48.8 & 180.6 & 87.7 \\ 
    \rowcolor{mygray}
    \textbf{Ours} & \textbf{22.1} & \textbf{109.6} & \textbf{77.9} \\ 
    \hline 
\end{tabular}}
\caption{Computation overhead comparison with the recent state-of-the-art methods that directly regress \textit{all 3D vertices}. The rightmost column shows the MPVE errors on the 3DPW test set for performance reference.}
\label{tab:computation}
\end{table}

\begin{table}[t]
\center
\small
\setlength{\tabcolsep}{4pt}
\resizebox{3.2in}{!}{
\begin{tabular}{l | c c c c c}
    \hline
     & \textbf{Ours} & \textit{w/o} $\mathcal{L}_{conf}$ & \textit{w/o} $\mathcal{L}_{pose}$ & \textit{w/o} $\mathcal{L}_{normal}$ & \textit{w/o} $\mathcal{L}_{edge}$ \\
    \hline
    MPVE$\downarrow$ & \textbf{58.0} & 59.2 & 58.3 & 60.6 & 60.4 \\
    
    \hline 
\end{tabular}}
\caption{MPVE errors on H3.6M \cite{h36m_pami} test set when ablating different loss terms.}
\label{tab:ablation_loss}
\end{table}

\begin{table}
    \centering
    \setlength{\tabcolsep}{12pt}
    \resizebox{3.2in}{!}{
    \begin{tabular}{l | c c c}
    \hline
    \multirow{2}{*}{Occ. VM Parts} & \multirow{2}{*}{MPVE$\downarrow$} & \multirow{2}{*}{MPJPE$\downarrow$} & \multirow{2}{*}{PA-MPJPE$\downarrow$} \\
    &  &  &  \\
    \hline
    \rowcolor{mygray}
    None (Ours) & \textbf{77.9} & \textbf{67.5} & \textbf{41.3} \\
    2 Arms & 79.2 \diff{$\uparrow$ 1.3} & 68.2 \diff{$\uparrow$ 0.7} & 42.2 \diff{$\uparrow$ 0.9} \\
    2 Legs & 78.3 \diff{$\uparrow$ 0.4} & 67.9 \diff{$\uparrow$ 0.4} & 41.7 \diff{$\uparrow$ 0.4} \\
    Body & 78.6 \diff{$\uparrow$ 0.7} & 68.0 \diff{$\uparrow$ 0.5} & 41.8 \diff{$\uparrow$ 0.5} \\
    Random & 78.7 \diff{$\uparrow$ 0.8} & 68.1 \diff{$\uparrow$ 0.6} & 41.9 \diff{$\uparrow$ 0.6} \\
    
    \hline 
\end{tabular}}
\caption{Results on 3DPW \cite{vonMarcard2018} test set when different parts of virtual markers (VM) are occluded.}
\label{tab:occl}
\end{table}

\subsubsection*{B.3 \quad Additional Quantitative Results}

\begin{figure*}[h]
    \centering
    \includegraphics[width=\linewidth]{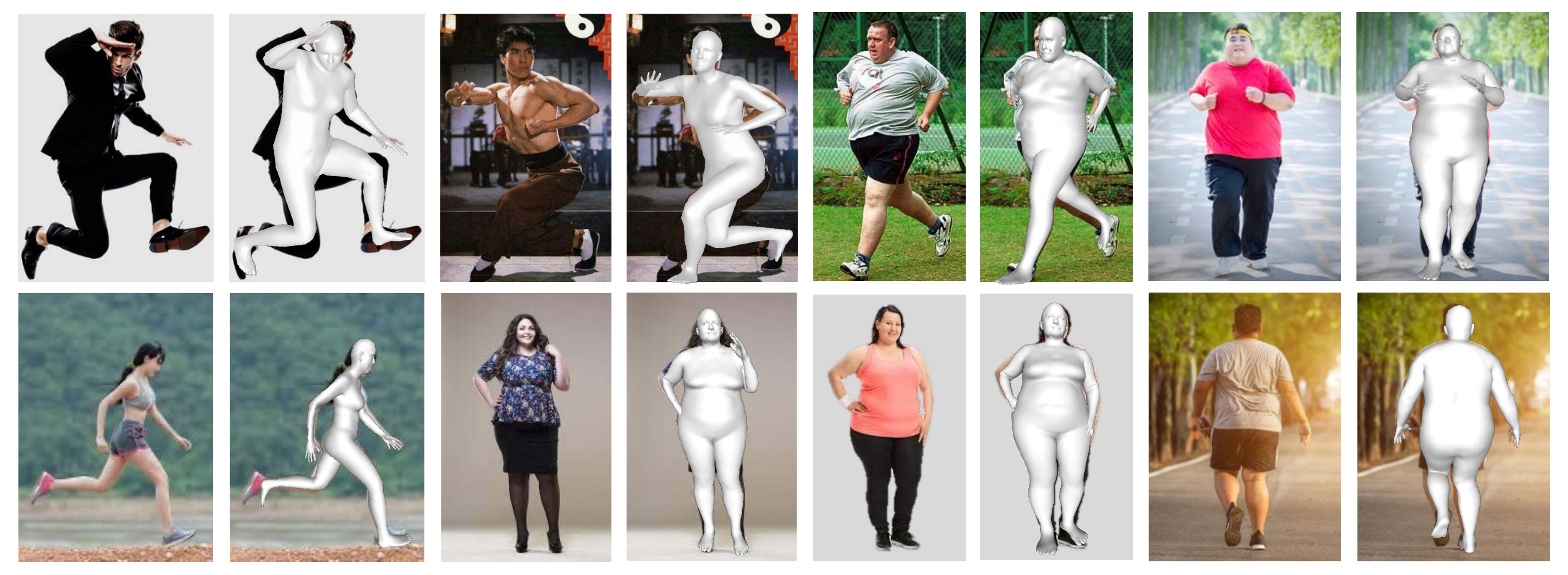}
    \caption{Meshes estimated by our approach on Internet images with challenging cases (complex poses or extreme body shapes).}
    \label{fig:supp_wild}
\end{figure*}

\paragraph{Different loss terms.}
Table \ref{tab:ablation_loss} reports the MPVE error on H3.6M \cite{h36m_pami} test set when ablating different loss terms. The confidence loss \cite{iskakov2019learnable} is used to encourage the interpretability of the heatmaps to have a maxima response at the GT position. Without the confidence loss, the error increases slightly. If ablating the surface losses, MPVE increases a lot, which demonstrates the smoothing effect of these two terms. 

\paragraph{Robustness to occlusion.}
We report results when different virtual markers are occluded by a synthetic mask in Table \ref{tab:occl}. The errors are slightly larger than the original image (None), which validates the effectiveness of the \textit{locality} of the virtual marker representation. Occluding arm regions results in a larger error increase. This may be because the arm has larger variations in the dataset. 

\paragraph{Comparison to fitting.}
In order to disentangle the ability of mesh regression from markers using $\hat{\mathbf{A}}$ with the ability to detect the virtual markers accurately from images, we first compute the estimation errors of the virtual markers. The MPJPE over all the virtual markers is $35.5$mm, which demonstrates that these virtual markers can be accurately detected from the images. 
We then fit the mesh model parameters to these virtual markers.
Table \ref{tab:fitting} shows the metrics of the fitted mesh on the SURREAL \cite{varol2017learning} test set. As we can see, the fitted mesh has a similar error as our regression ones which uses the interpolation matrix $\hat{\mathbf{A}}$, which validates the accuracy of the estimated virtual markers.

\subsubsection*{B.4 \quad Additional Qualitative Results}
\begin{figure*}[t]
    \centering
    \includegraphics[width=5.7in]{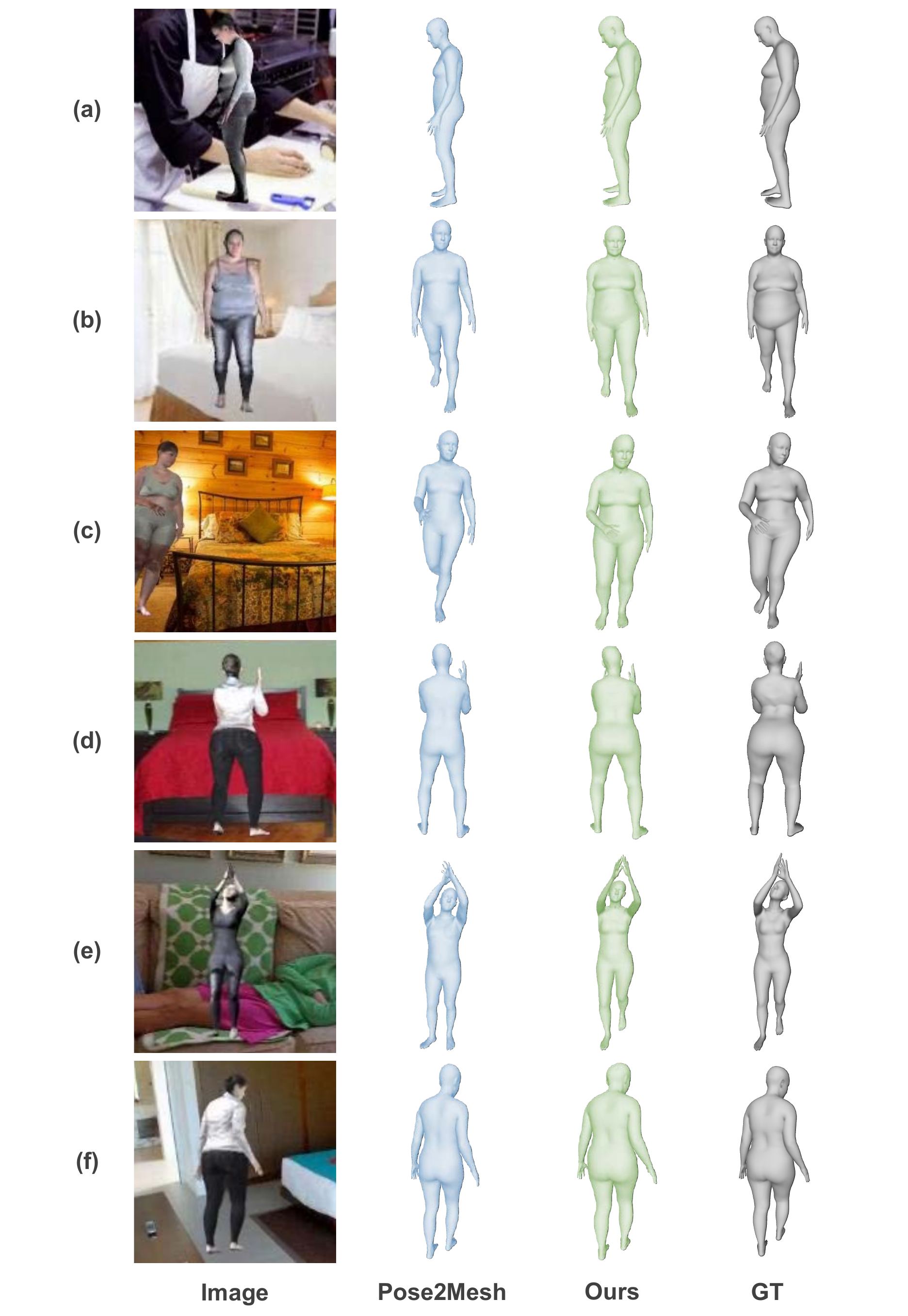}
    \caption{Qualitative comparison between our method and Pose2Mesh \cite{choi2020pose2mesh} on SURREAL test set \cite{varol2017learning}. Our approach generates more accurate mesh estimation results on images of diverse body shapes.}
    \label{fig:supp_surreal}
\end{figure*}

Figure \ref{fig:supp_surreal} shows more qualitative comparisons with Pose2Mesh \cite{choi2020pose2mesh} on the SURREAL test set in which has diverse body shapes. The skeleton representation used in Pose2Mesh loses the body shape information so the method \cite{choi2020pose2mesh} can only recover mean shapes. For example, in Figure \ref{fig:supp_surreal} (d) (e), the estimated meshes of Pose2Mesh tend to have the average body shape and fail to estimate the real body shape, regardless of whether the person is slim or stout. This is caused by the limited skeleton representation bottleneck so that the model learns a mean shape for the whole training dataset implicitly. In contrast, our approach with virtual marker representation generates more accurate mesh estimation results.  

\begin{table}
    \centering
    \setlength{\tabcolsep}{16pt}
    \resizebox{3.2in}{!}{
    \begin{tabular}{l | c c c}
    \hline
    \multirow{2}{*}{Method} & \multirow{2}{*}{MPVE$\downarrow$} & \multirow{2}{*}{MPJPE$\downarrow$} & \multirow{2}{*}{PA-MPJPE$\downarrow$} \\
    &  &  &  \\
    \hline
    Fitting & 44.6 & 34.8 & 29.5  \\
    \rowcolor{mygray}
    Ours & 44.7 & 36.9 & 28.9 \\
    
    \hline 
\end{tabular}}
\caption{Results on SURREAL \cite{varol2017learning} test set when the mesh is obtained by fitting to the estimated virtual markers.}
\label{tab:fitting}
\end{table}

\begin{figure*}[t]
    \centering
    \includegraphics[width=6.6in]{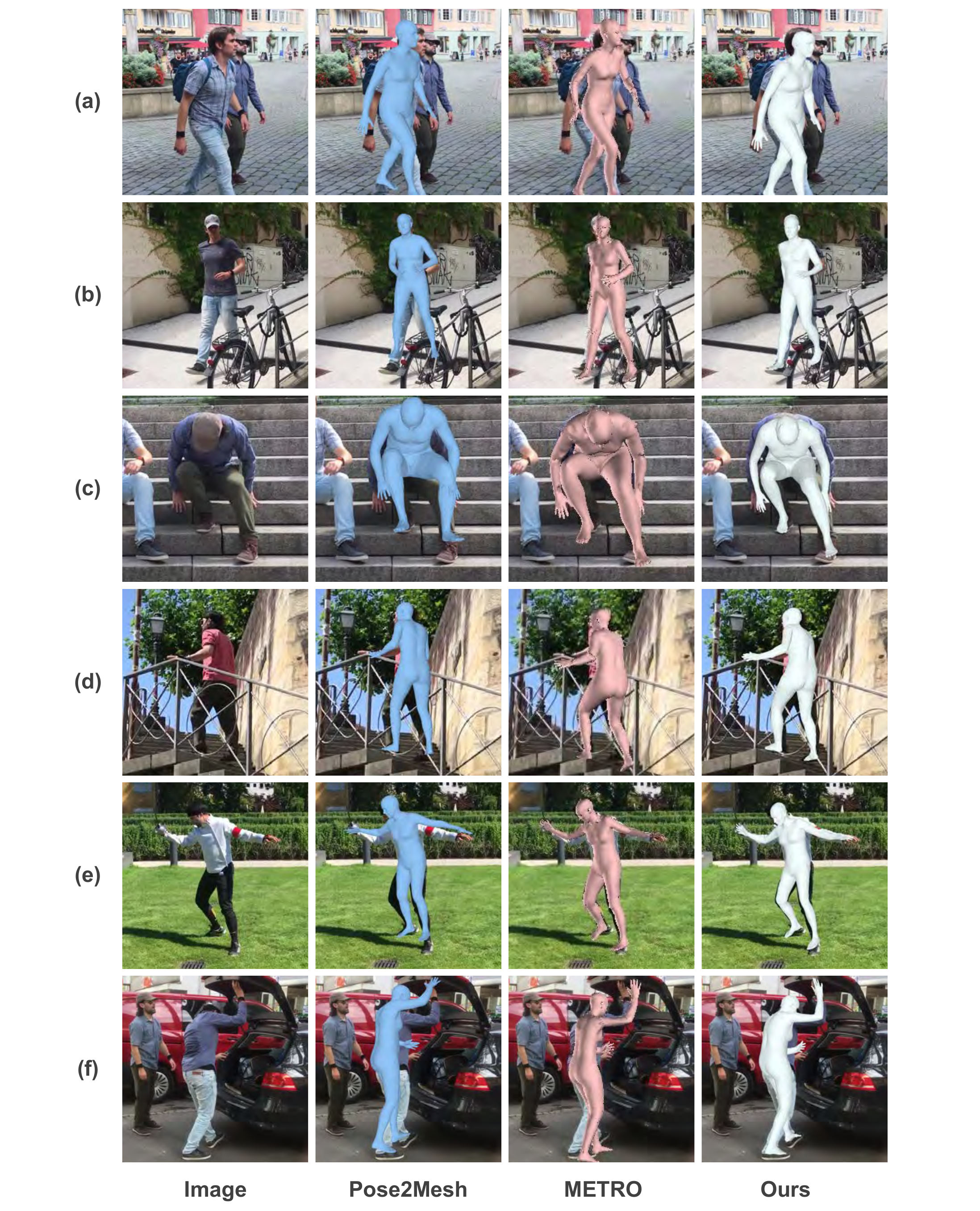}
    \caption{Qualitative comparison between our method and Pose2Mesh \cite{choi2020pose2mesh}, METRO \cite{lin2021end} on 3DPW test set \cite{vonMarcard2018}. Our approach is more robust to occlusion and truncation and generates more accurate mesh estimation results that align images well.}
    \label{fig:supp_pw3d}
\end{figure*}

Figure \ref{fig:supp_pw3d} shows more qualitative comparisons with Pose2Mesh \cite{choi2020pose2mesh} and METRO \cite{lin2021end} on the 3DPW test set. Pose2Mesh and METRO use the skeleton or all 3D vertices as intermediate representations, respectively. The estimated meshes are overlaid on the images according to the camera parameters. Pose2Mesh \cite{choi2020pose2mesh} has difficulty in estimating correct body pose and shapes when truncation occurs (a) or in complex postures (c). The results of METRO \cite{lin2021end} have many artifacts where the estimated mesh is not smooth, and they also fail to align the image well.  In contrast, our method estimates more accurate human poses and shapes and has smooth human mesh results. In addition, it is more robust to truncation and occlusion and aligns the image better.

\begin{figure}[t]
    \centering
    \includegraphics[width=3.3in]{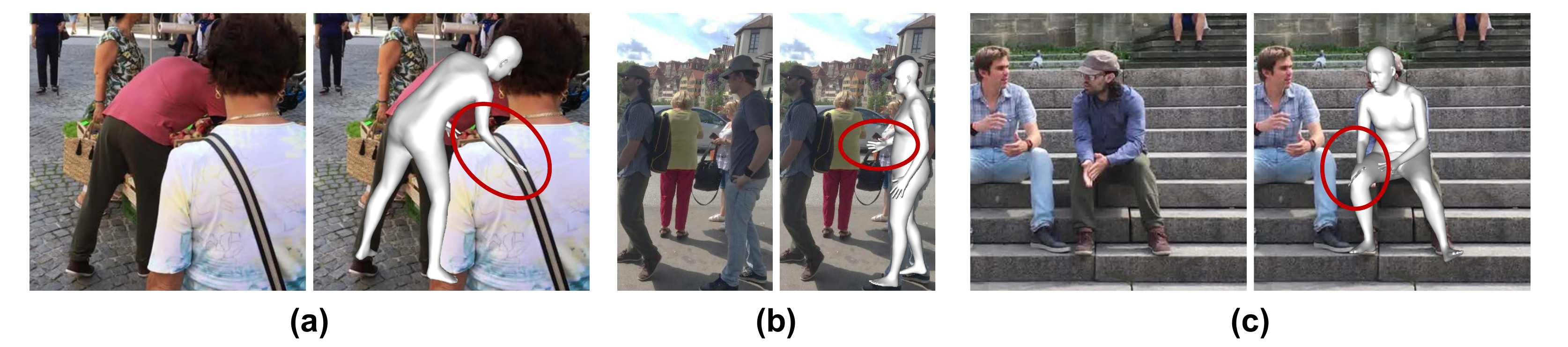}
    \caption{Typical failure cases. \textbf{(a)} The right arm has inaccurate shape estimation due to the inaccurate virtual marker estimation around the arm when occluded. \textbf{(b)} Our method treats the lower arm of another person as its own due to occlusion. \textbf{(c)} Interpenetration around the right hand.}
    \label{fig:supp_failure}
\end{figure}

\begin{figure*}[t]
    \centering
    \includegraphics[width=6.5in]{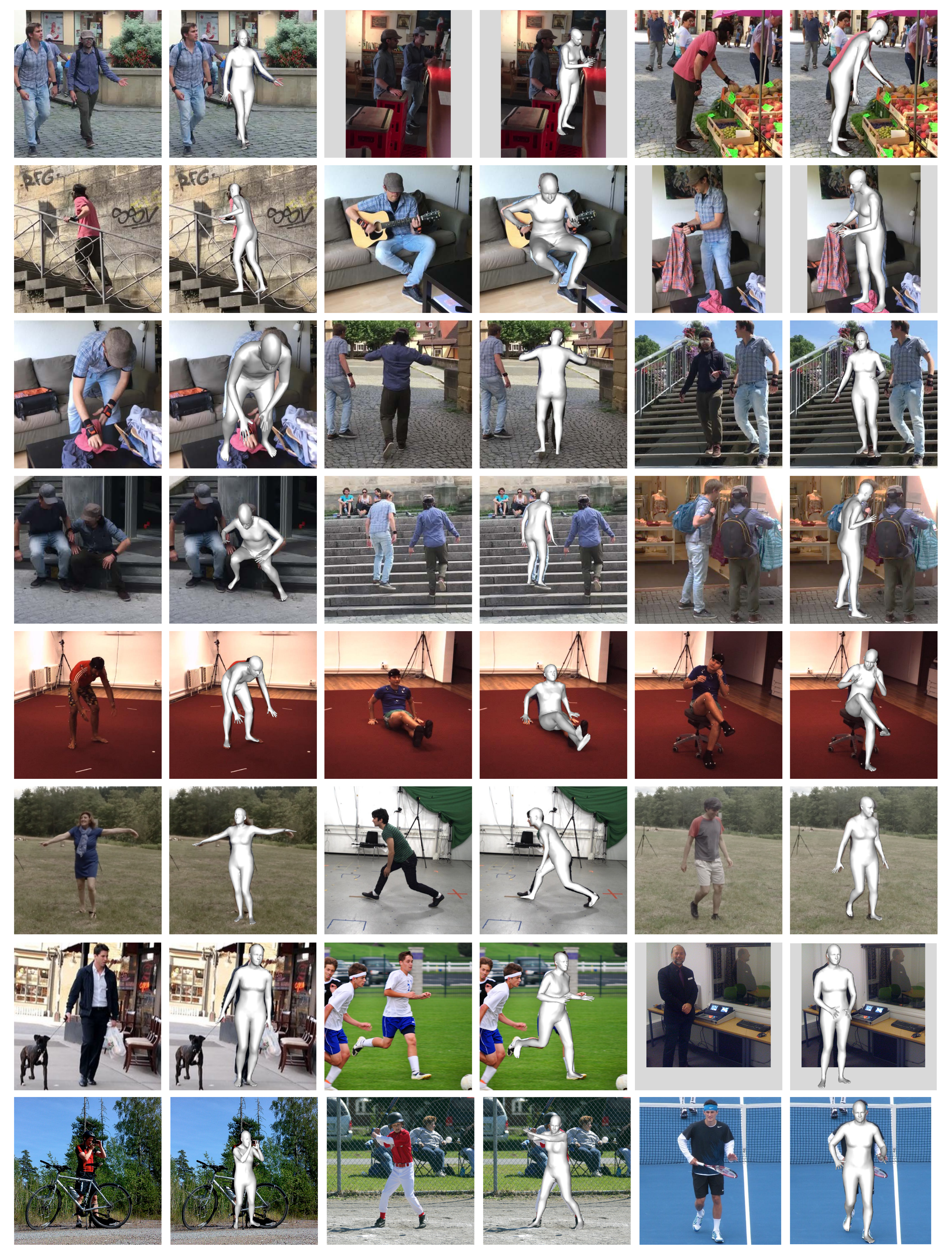}
    \caption{Meshes estimated by our approach on images from the 3DPW \cite{vonMarcard2018} dataset (row 1-4), H3.6M \cite{h36m_pami} dataset (row 5), MPI-INF-3DHP \cite{mehta2017monocular} dataset (row 6), and COCO dataset (last 2 rows) \cite{lin2014microsoft}. }
    \label{fig:supp_quality_results}
\end{figure*}

Figure \ref{fig:supp_quality_results} shows more quality results of our approach on the 3DPW \cite{vonMarcard2018}, H3.6M \cite{h36m_pami}, MPI-INF-3DHP \cite{mehta2017monocular}, and COCO \cite{lin2014microsoft} datasets. Figure \ref{fig:supp_wild} shows more qualitative results on Internet images with challenging cases, such as extreme body shapes or complex poses. Our method generalizes well on the natural scenes. Figure \ref{fig:supp_failure} shows typical failure cases, including inaccurate shape estimation and interpenetration, which are mainly caused by inaccurate 3D virtual marker estimation when occlusion occurs. But as expected, the rest body parts are barely affected due to the \emph{local and sparse} property of the virtual marker.

\subsection*{C. \quad Human Subject Data} 

We use existing public datasets of human subjects in our experiments following their official licensing requirements. With proper usage, the proposed method could be beneficial to society (\eg elderly fall detection).

\end{document}